\documentclass{article}
\usepackage{ijcai24}
\usepackage{commands}

\usepackage{makecell}
\usepackage{times}
\usepackage{soul}
\usepackage{url}
\usepackage[hidelinks]{hyperref}
\usepackage[utf8]{inputenc}
\usepackage[small]{caption}
\usepackage{graphicx}
\usepackage{amsmath}
\usepackage{amsthm}
\usepackage{booktabs}
\usepackage[ruled]{algorithm2e}
\SetAlFnt{\small}
\SetAlCapFnt{\small}
\SetAlCapNameFnt{\small}
\usepackage[switch]{lineno}
\usepackage{subcaption}
\usepackage{bbm}
\usepackage{multirow}
\usepackage[table,xcdraw]{xcolor}
\usepackage{color,soul}

\title{\algorithmname{}: Random Effects based Online RL algorithm for Reducing Cannabis Use}

\author{
Susobhan Ghosh$^1$
\and
Yongyi Guo$^2$\and
Pei-Yao Hung$^{3}$\and
Lara Coughlin$^4$\and
Erin Bonar$^4$\and\\
Inbal Nahum-Shani$^3$\and
Maureen Walton$^4$\and
Susan Murphy$^1$
\\
\affiliations
\small
$^1$Department of Computer Science, Harvard University\\
$^2$Department of Statistics, University of Wisconsin-Madison\\
$^3$Institute for Social Research, University of Michigan\\
$^4$Department of Psychiatry, University of Michigan\\
\emails
\small
susobhan\_ghosh@g.harvard.edu,
guo98@wisc.edu,
peiyaoh@umich.edu,
laraco@med.umich.edu,
erinbona@med.umich.edu,
inbal@umich.edu,
waltonma@med.umich.edu,
samurphy@fas.harvard.edu
}

\begin{document}

\maketitle

\begin{abstract}
    The escalating prevalence of cannabis use, and associated cannabis-use disorder (CUD), poses a significant public health challenge globally. With a notably wide treatment gap, especially among emerging adults (EAs; ages 18-25), addressing cannabis use and CUD remains a pivotal objective within the 2030 United Nations Agenda for Sustainable Development Goals (SDG). In this work, we develop an online reinforcement learning (RL) algorithm called \algorithmname{} which will be utilized in a mobile health study to deliver personalized mobile health interventions aimed at reducing cannabis use among EAs. \algorithmname{} utilizes \emph{random effects} and \emph{informative Bayesian priors} to learn quickly and efficiently in noisy mobile health environments. Moreover, \algorithmname{} employs Empirical Bayes and optimization techniques to autonomously update its hyper-parameters online.
    To evaluate the performance of our algorithm, we construct a simulation testbed using data from a prior study, and compare against commonly used algorithms in mobile health studies. We show that \algorithmname{} performs equally well or better than all the baseline algorithms, and the performance gap widens as population heterogeneity increases in the simulation environment, proving its adeptness to adapt to diverse population of study participants.
\end{abstract}

\section{Introduction \& Motivation}
\label{sec:intro}

Addressing at-risk substance use, including cannabis use, is a pivotal objective within the 2030 UN Agenda for Sustainable Development Goals (SDG)\footnote{\label{fn2}https://sdgs.un.org/2030agenda}. Within this agenda, SDG 3 focuses on ensuring healthy lives and well-being across the lifespan, yet, increasing use of cannabis, third in global prevalence after alcohol and nicotine, threatens this goal \cite{peacock2018global}. Hence, as highlighted in target 3.5 of the agenda, strengthening the prevention and treatment of cannabis use and cannabis use disorder (CUD) is crucial. Unfortunately, this coincides with a decreased public perception of the risks associated with cannabis use, likely influenced by ongoing decriminalization efforts and greater access to cannabis products \cite{carliner2017cannabis}, further worsened by one of the largest treatment gaps of any medical condition, with one study showing only ~5\% of those with CUD receiving treatment \cite{lapham2019prevalence}. 

In the US, the prevalence of cannabis use is highest among emerging adults (EAs; age 18-25) \cite{SAMHSA}, marking it as a significant concern within the growing landscape of cannabis use. Particularly worrisome is the fact that early initiation of cannabis use links to an array of physical and mental health repercussions, as well as escalated risk for developing CUD \cite{volkow2014adverse,hall2009adverse,chan2021young,hasin2016prevalence}. Given that cannabis use frequently commences during adolescence and peaks in emerging adulthood, this is a critical developmental period for early intervention strategies to prevent transitions into CUD.

Mobile health technologies, such as health apps and sensors, can potentially serve as support tools to help individuals manage their cannabis use. Using these tools, individuals can track their cannabis consumption, receive personalized interventions, and provide objective data for early detection of issues. These technologies enable the delivery of just-in-time adaptive interventions (JITAIs) \cite{nahum2018just}, which leverage rapidly changing information about a person's state and context to decide whether and how to intervene in daily life. JITAIs have been successful for many domains of behavioral health \cite{jaimes2015preventer,clarke2017mstress,golbus2021microrandomized}, whilst JITAIs for cannabis use among EAs are currently lacking evidence despite promising early data \cite{shrier2018pilot}. 

In this work, we develop an RL algorithm called \algorithmname{} which will be utilized in the \trialname pilot study (Section \ref{sec:miwaves}). \trialname focuses on developing a JITAI for reducing cannabis use among emerging adults (EAs) (ages 18-25). This JITAI  leverages \algorithmname{} to determine the likelihood of delivering an intervention message.

\subsection{\trialname pilot study}
\label{sec:miwaves}
\begin{figure*}[!t]
    \centering
    \includegraphics[width=0.7\linewidth]{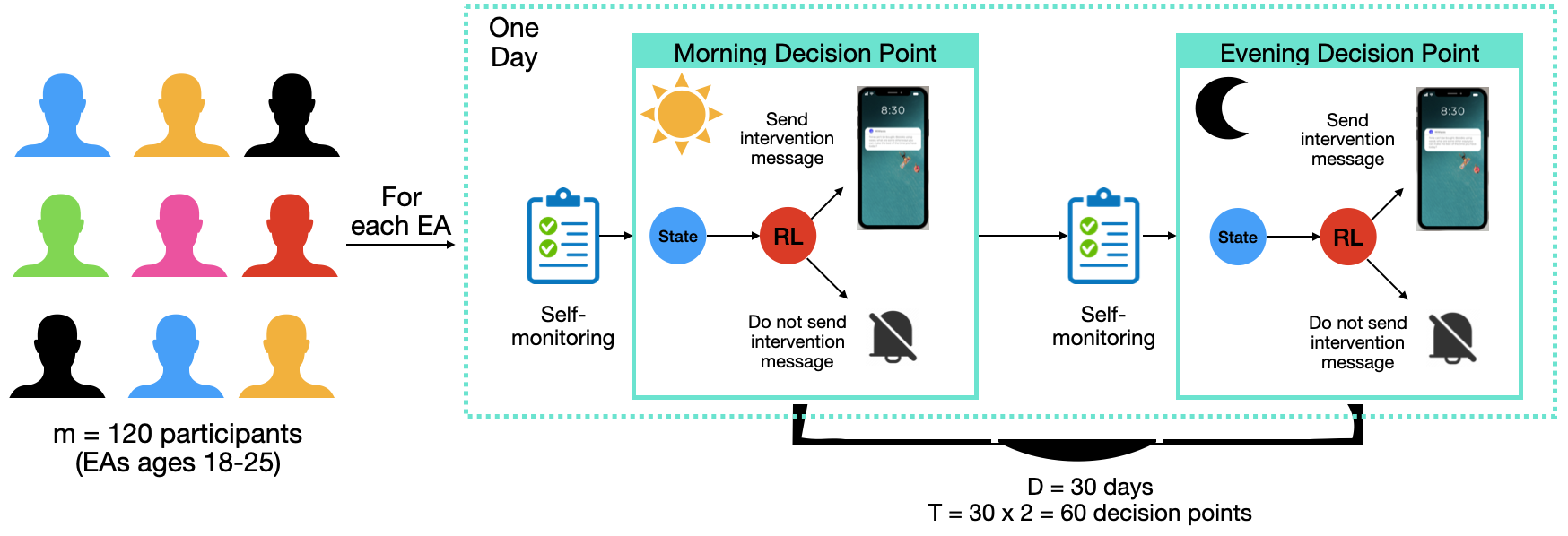}
    \caption{Summary of the \trialname pilot study. $m=120$ EAs are expected to be recruited through social media ads. Each EA will be in the trial for 30 days, and will be asked to self-report twice daily - once in the morning and once in the evening. Upon completion or time expiration of the self-reporting, the RL algorithm will decide whether to send or not send an intervention message.}
    \label{fig:miwaves}
\end{figure*}

The \trialname pilot study focuses on developing a \emph{personalizing} Just-In-Time Adaptive Intervention (pJITAI), namely a JITAI that integrates an RL algorithm. In this study, EAs are randomized to receive a mobile-based intervention message
or no message, twice daily. The RL algorithm is designed to learn from a participant's history, and \emph{personalize} the likelihood of intervention delivery based on a participant's current context. By combining technology, behavioral science, and data-driven decision-making, \trialname aims to empower emerging adults with the digital tools to help reduce their cannabis use. The \trialname pilot study has been registered on ClinicalTrials.gov (NCT05824754), and is scheduled to start in \bo{March 2024}. Figure \ref{fig:miwaves} provides a visual overview of the \trialname pilot study.

\subsection{Challenges, Contributions and Overview}
Deploying RL algorithms in mHealth studies like \trialname present a multitude of challenges that must be addressed, which include:
\begin{enumerate}[label=\textbf{C\arabic*}]
    \item \label{c1} \bo{Limited Data}: Many sequential decision making problems in mHealth involve scarce data, forcing RL algorithms to learn and perform well under strict data constraints \cite{trella2022designing}.
    \item \label{c2} \bo{After-study analysis and evaluation:} The RL algorithms deployed in mHealth studies need to be developed in a way to facilitate after-study analysis and off-policy evaluation.
    \item \label{c3} \bo{Autonomy and Stability}: 
    The intervention protocol in clinical studies is pre-specified. Since the RL algorithm is part of the intervention, scientists do not have the flexibility to change the RL algorithm while the study is running.
    RL algorithms must exhibit robustness in the face of noisy data, ensuring consistent and reliable performance throughout the study \cite{trella2022designing}.
    \item \label{c4} \bo{Explainability}: It is imperative that RL algorithms are interpretable and comprehensible to behavioral scientists and medical professionals to enhance their ability to critique RL performance and to enhance the possibility of larger scale implementation.  
    \item \label{c6} \bo{Delayed Effects}: In mobile health studies, each intervention message sent to the user has a \emph{delayed effect}. Users may perceive burden upon receiving an intervention message, which influences their future behavior.
    \item \label{c5} \bo{Reproducibility}: Any algorithm used as part of the intervention in a clinical study needs to be reproducible in order for health scientists to evaluate and implement the intervention package in practice. Hence, the decisions taken by the RL algorithm must be reproducible, allowing for scrutiny and verification of their effectiveness. 
\end{enumerate}

To that end, we introduce \algorithmname, an online RL algorithm which utilizes \emph{random effects} to address the challenges mentioned above.
When used as part of an RL algorithm, random effects allow the algorithm to learn quickly and efficiently by making use of other participant's data in the population while simultaneously personalizing treatment for a given participant. Moreover, \algorithmname{} employs an informative Bayesian prior formulated from pre-existing data to act as a warm-start. Carefully designed priors incorporate previous (domain) knowledge, which help algorithms learn quickly and efficiently. Both \emph{random effects} and informative \emph{priors} can help \algorithmname{} to tackle challenge \ref{c1}.

The most commonly used RL algorithms in mHealth settings are \emph{bandit} algorithms. In mHealth settings, predictions of the value of next state (eg. using \cite{jiang2015dependence}) can be very noisy. Bandit algorithms are, thus, preferred due to their performance in such noisy environments.
Moreover, they are computationally less complex, and hence are able to run stably and reliably in an online environment. 
Further, linear models are often considered interpretable due to their simplicity of representing the role of various factors, and can also be stably updated. \algorithmname{} utilizes both these concepts - it uses a bandit framework, along with a linear model (with random effects) to model the reward. We derive the formula to update \algorithmname's parameters and hyper-parameters online (Sec \ref{sec:rl_update}). 
We show that we are able to autonomously update these parameters and hyper-parameters within a reasonable time-limit in an online environment. Moreover, to facilitate after-study analysis, we utilize a smooth variant of posterior sampling, and clip the probabilities of taking an action (Sec \ref{sec:act_select}). This way, \algorithmname{} is able to overcome challenges \ref{c2}, \ref{c3} and \ref{c4}.




To address delayed effects (\ref{c6}), one can use RL algorithms to model a full Markov Decision Process (MDP). However, in mobile health settings, such approaches are not feasible due to limited data and noisy outcomes \cite{trella2023reward}. On the other hand, the classical bandit framework alone is also insufficient, as it is designed to optimize for immediate reward, and thus, cannot account for the delayed effects of actions. To that end, we engineer the reward used to update \algorithmname{}'s parameters and hyper-parameters (Sec \ref{sec:reward_design}), to account for delayed effects of actions.

Finally, to tackle challenge \ref{c5}, we have made our implementation of \algorithmname{} publicly available\footnote{\url{https://github.com/StatisticalReinforcementLearningLab/miwaves_rl_service}
}.
To ensure reproducibility, we employ a seeded pseudo-random number generator to make every stochastic decision in \algorithmname{}. Additionally, all intermediate results and values used to make decisions are programatically stored in a database for reproducing the results obtained during any \emph{run} of the algorithm.

\section{RL Framework and Notation}
\label{sec:rl_framework}
This section provides a brief overview of the Reinforcement Learning (RL) \cite{sutton2018reinforcement} setup used in this work, and the specifics of the RL setup with respect to the \trialname pilot study.

We approximate the pilot study environment as a bandit environment. We represent it as a Markov Decision Process (MDP) 
where in the RL algorithm (eg. the mobile app) interacts with the environment (eg. the user). The MDP is specified by the tuple $\langle \mathcal{S}, \mathcal{A}, r, P, T \rangle$, where $\mathcal{S}$ is the state-space of the algorithm, $\mathcal{A}$ is the action-space, $r(s,a)$ is the reward function defined for a given state $s \in \mathcal{S}$ and action $a \in \mathcal{A}$, $P(s, a, s')$ is the transition function for a given state $s \in \mathcal{S}$, action $a \in \mathcal{A}$ and next state $s' \in \mathcal{S}$, and $T$ is the total number of decision times. A user trajectory is given by $\mathcal{H}^{(T+1)}_{i} = \{ S^{(t)}, \action{t}{}, R^{(t)} \}_{t= 1}^{T}$, where $S^{(t)}$ denotes the state at decision time $t$, $\action{t}{}$ the action assigned by the RL algorithm at time $t$, and $R^{(t)}$ the reward collected after selection of the action. In the case of \trialname we have:\\
\textbf{Actions}: Binary action space, i.e. $\mathcal{A} = \{0, 1\}$ - to not send ($0$) or to send ($1$) an intervention message.\\
\textbf{Decision points}: $T$ decision points per user. The study is set to run for $D = 30$ days, and each day is supposed to have 2 decision points per day. Therefore, we expect to have 60 decision points per user, i.e. $T = 60$.\\
\textbf{Reward}: We denote the reward for user $i$ at decision time $t$ by $R_{i}^{(t)}$. For the \trialname pilot study, we have discrete rewards $\{0, 1, 2, 3\}$, which increase linearly with user engagement. We utilize engagement as our reward because engagement is critical to assess effectiveness of interventions after the study is over \cite{nahum2022engagement}.\\
\textbf{States}: Let us denote the state observation of the user $i$ at decision time $t$ as $S_{i}^{(t)}$. A given state $S = (S_1, S_2, S_3)$
is defined as a 3-tuple of the following binary variables (omitting the user and time index for brevity):
    \begin{itemize}
        \item $S_1$: Recent engagement - set to $1$ if the average of past $3$ observed rewards is greater than or equal to $2$ (high engagement), and set to $0$ otherwise (low engagement).
        At decision point $t=1$, we set $S_1$ to 0, as there is no engagement by the user at the start of the pilot study.
        
        \item $S_2$: Time of day of the decision point - Morning (0) vs. Evening (1).

        \item $S_3$: Recent cannabis use - set to $0$ if the participant reported using cannabis during their self-monitoring, and $1$ otherwise.
        If the user fails to self-report, we set $S_3$ to be 0. We do so because we expect the participant in the \trialname pilot study to be using cannabis regularly (at least 3 times a week).
    \end{itemize}
    Overall, we represent the favorable states as $1$ (not using cannabis, high engagement), and the unfavorable states as $0$ (using cannabis, low engagement).\\
\textbf{Number of users}: We expect $m=120$ users to participate during the RL-powered \trialname pilot study.


\section{Related Work}
\label{sec:related_work}
Random effects (and mixed-effects) models have been well-studied in the statistical literature \cite{laird2004random,laird1982random,robinson1991blup}, mainly in the context of batch data analysis. Mixed-effects models comprise of fixed and \emph{random effects} - hence termed mixed effects. \citeauthor{laird1982random} introduce the notion of random-effects models for longitudinal data, and describe an unified approach to fitting such models using empirical Bayes and maximum likelihood estimation using EM algorithm. Our work (which is in context of streaming / real-time data) draws inspiration from \citeauthor{laird1982random} to extend random-effects based models to real-time decision making through RL in sequential decision making problems.

There has been a myriad of works in optimizing intervention delivery in mHealth settings in recent years \cite{golbus2021microrandomized,kramer2019investigating,martin2018development,rabbi2019optimizing,trella2022designing,walsh2019integrating}. 
\emph{Bandit} algorithms are the most commonly used RL algorithms used in such high stakes online settings \cite{langford2007epoch,tewari2017ads,wang2005bandit} due to their simplicity and stability, and ability to perform in noisy environments. 
Such algorithms have mainly used one of two approaches. The first approach is person specific (a.k.a. fully personalized) \cite{forman2019can,jaimes2015preventer,peng2019,rabbi2015mybehavior} where a separate model is deployed for each user in the trial. This approach is suitable when the population of users are highly heterogeneous, but suffers greatly when data is scarce and/or noisy. Note that fully personalized approaches are not feasible for the \trialname pilot study, as it runs for only $30$ days (due to scarce data). The second approach completely pools data (a.k.a. fully pooled) across all users in the population \cite{clarke2017mstress,paredes2014poptherapy,trella2022designing,yom2017encouraging,zhou2018personalizing}. Our algorithm, \algorithmname{}, strikes a balance between the two approaches - it adaptively pools data across users depending on the degree of heterogeneity in the population. \ref{sec:reward_model} describes in detail as to how we achieve that balance using \emph{random effects}.

\citeauthor{tomkins2021intelligentpooling} also use random effects in their Thompson-Sampling~\cite{russo2014learning,thompson1933likelihood} contextual bandit algorithm \cite{li2010contextual}, IntelligentPooling~\cite{tomkins2021intelligentpooling}.  IntelligentPooling updates its hyper-parameters by modeling the problem as a Gaussian Process (GP). However,
IntelligentPooling fails to run autonomously and stably in an online environment
(does not overcome \ref{c3})\footnote{We were unable to run their code published on GitHub, and unable to parse the code or replicate it due to poor documentation}. 
Here we deal with this problem by updating the hyper-parameters in \algorithmname{} using empirical Bayes (similar to \cite{laird1982random}), and solve the optimization problem using projected gradient descent. \algorithmname{} runs autonomously and stably in an online environment while also having more users (12x) and more features (8x) as compared to the environment described in IntelligentPooling.

Recently, various approaches have been explored regarding the application of mixed effects models within a bandit framework \cite{zhu2022random,zhu2022robust,aouali2023mixed}. However, these works primarily focus on utilizing mixed effects to capture the dependence and heterogeneity of rewards associated with different actions, rather than addressing the similarity and heterogeneity among multiple users. For example, \cite{zhu2022random} consider a (non-contextual) multi-arm bandit problem, where the agent chooses one of the $K$ arms at each time $t$ with the goal of maximizing cumulative reward. The authors assume that the reward for the arms are correlated with each other and can be expressed using a mixed-effects model, so that pulling an arm gives some information about the reward of other arms as well. A follow up work, \cite{zhu2022robust}, adds context into the reward model, and results in a linear mixed effects model for the rewards of the arms. \cite{aouali2023mixed} further generalizes the above works to a non-linear reward setting. Our approach diverges from these studies by utilizing mixed-effects to model user similarity and heterogeneity, while making decisions for each user at each time point.

In the broader RL literature, there has been much work on Thompson Sampling based bandit algorithms \cite{basu2021no,hong2022hierarchical}, especially in connection to multi-task learning and meta learning \cite{peleg2022metalearning,simchowitz2021bayesian,wan2021metadata,wan2023towards}. The multi-task learning based approaches quantify the similarity between arms and/or users from their policies - the extent to which one user's data influences or contributes to another user's policy is a function of some similarity measure. 
\algorithmname{} can be connected to multi-task learning, as it learns across multiple users (or tasks), and tries to maximize rewards across all users (or tasks). However, due to its unique application in mobile health, reBandit adopts a distinct set of assumptions on the structure of the similarity measures in comparison to the works mentioned above.
The meta-learning based approaches exploit the underlying structure of similar tasks to improve performance on new (or unseen, but similar) tasks. While \algorithmname{} can be viewed as a form of meta-learning, where shared population parameters are learnt across users (or tasks), and user-specific parameters are learnt to personalize to tasks, \algorithmname{} does not try to improve performance on new or unseen users.

\section{Bandit algorithm: \algorithmname{}}
\label{sec:rl_alg}
This section details details the \algorithmname{} algorithm used in the \trialname{} pilot study. Being an online RL algorithm, \algorithmname{} has two major components: (i) the online learning algorithm; and (ii) the action-selection procedure. Going forward, we describe \algorithmname{}'s online learning algorithm in Section \ref{sec:online_rl_alg}, and it's posterior sampling based action selection strategy in Section \ref{sec:act_select}. Finally, to address delayed effects, we describe its reward engineering procedure in Section \ref{sec:reward_design}. The \algorithmname{} algorithm is summarized in Algorithm \ref{algo:multistate_mixed_algo}.

\subsection{Online Learning Algorithm}
\label{sec:online_rl_alg}

This section details the online learning algorithm - specifically the algorithm's reward approximating function and its model update procedure.

\subsubsection{Reward Approximating Function}
\label{sec:reward_model}
One of the key components of the online learning algorithm is its reward approximation function, through which it models the participant's reward. Recall that the reward function is the conditional mean of the reward given state and action. We chose a Bayesian Mixed Linear Model to model the reward.
Mixed models allow the RL algorithm to adaptively pool and learn across users while simultaneously personalizing actions for each user.  

Let us assume that for a given user $i$ at decision time $t$, the RL algorithm receives the reward $\reward{t}{i}$ after taking action $\action{t}{i}$ 
Then, the reward model is written as:
\begin{align}
    R_{i}^{(t)} 
    = g(\state{t}{i})^T \boldsymbol{\alpha_i}  + \action{t}{i} f(\state{t}{i})^T \boldsymbol{\beta_i} + \epsilon_i^{(t)}
\end{align}
where $\epsilon_i^{(t)}$ is the noise, assumed to be gaussian i.e. $\bs{\epsilon} \sim \mathcal{N}(\bs{0}, \sigma_{\epsilon}^2\bs{I}_{mt})$, and $m$ is the total number of users who have been or are currently part of the study. Also $\boldsymbol{\alpha_i}$, $\boldsymbol{\beta_i}$, and $\boldsymbol{\gamma_i}$ are weights that the algorithm wants to learn. $g(S)$ and $f(S)$ are functions of the RL state defined in Section \ref{sec:rl_framework}.
To enhance robustness to misspecification of the baseline reward model when $\action{t}{i}=0$, $g(\state{t}{i})^T \boldsymbol{\alpha_i} $, we utilize action-centering \cite{greenewald2017action} to learn an over-parameterized version of the above reward model:
\begin{align}
    R_{i}^{(t)} 
    &= g(\state{t}{i})^T \boldsymbol{\alpha_i}  + (\action{t}{i} - \pii{t}{i}) f(\state{t}{i})^T \boldsymbol{\beta_i} \nonumber \\
    & + (\pi_i^{(t)})f(\state{t}{i})^T \boldsymbol{\gamma_i} + \epsilon_i^{(t)}
    \label{eqn:reward_act_center}
\end{align}
where $\pii{t}{i}$ is the probability of taking action $\action{t}{i} = 1$ in state $\state{t}{i}$ for participant $i$ at decision time $t$. We refer to the term $g(\state{t}{i})^T \boldsymbol{\alpha_i}$ as the baseline, and $f(\state{t}{i})^T \boldsymbol{\beta_i}$ as the advantage (i.e. the advantage of taking action 1 over action 0).

We re-write the reward model as follows:
\begin{align}
    R_{i}^{(t)} 
    &= \Phii{T}{it} \tparam{}{i} + \epsilon_{i,t}
\end{align}
where $\Phii{T}{it} = \Phii{}{}(\state{t}{i}, \action{t}{i}, \pii{t}{i})^T = [g(\state{t}{i})^T, (\action{t}{i} - \pii{t}{i}) f(\state{t}{i})^T, (\pi_i^{(t)})f(\state{t}{i})^T]$ is the design matrix for given state and action, and $\tparam{}{i} = [\boldsymbol{\alpha_i}, \boldsymbol{\beta_i}, \boldsymbol{\gamma_i}]^T$ is the joint weight vector that the algorithm wants to learn. We further break down the joint weight vector $\tparam{}{i}$ into two components:
\begin{align}
    \tparam{}{i} = \begin{bmatrix}
            \boldsymbol{\alpha}_i\\
            \boldsymbol{\beta}_i\\
            \boldsymbol{\gamma}_i
        \end{bmatrix}
        = \begin{bmatrix}
        \boldsymbol{\alpha}_{\text{pop}} + \boldsymbol{u}_{\alpha, i}\\
        \boldsymbol{\beta}_{\text{pop}} + \boldsymbol{u}_{\beta, i}\\
        \boldsymbol{\gamma}_{\text{pop}} + \boldsymbol{u}_{\gamma, i}
        \end{bmatrix}
        = \tpop{} + \ui{}{i}
\end{align}
Here, $\tpop{}{} = [\boldsymbol{\alpha}_{\text{pop}}, \boldsymbol{\beta}_{\text{pop}}, \boldsymbol{\gamma}_{\text{pop}}]^T$ is the population level term which is common across all the user's reward models and follows a normal prior distribution given by $\tpop{} \sim \mathcal{N}(\muprior, \Sigprior)$. On the other hand, $\ui{}{i} = [\boldsymbol{u}_{\alpha, i}, \boldsymbol{u}_{\beta, i}, \boldsymbol{u}_{\gamma, i}]^T$ are the individual level parameters, or the \emph{random effects}, for any given user $i$. Note that the individual level parameters are assumed to be normal by definition, i.e. $\ui{}{i} \sim \mathcal{N}(\bs{0}, \Sig{}{u})$, and independent of $\epsilon_i$. Please refer to Appendix \ref{sec:priors} for the prior calculation and initialization values.



\subsubsection{Online model update procedure}
\label{sec:rl_update}


\textbf{Posterior Update}: We vectorize the terms across the $m$ users in a study, and re-write the model as:
{\allowdisplaybreaks
\begin{align}
\allowdisplaybreaks
    \bs{R} &= \Phii{T}{} \bs{\theta} +\bs{\epsilon}\\
    \bs{R}_i &=
    \begin{bmatrix}
        \reward{1}{i}
        \ldots
        \reward{t}{i}
    \end{bmatrix}^T\\
    \bs{R} &=
    \begin{bmatrix}
        \bs{R^T_{1}}
        \ldots
        \bs{R^T_{m}}
    \end{bmatrix}^T\\
    \bs{\theta} &= \begin{bmatrix}
        \bs{\theta}^T_1
        \ldots
        \bs{\theta}^T_m
    \end{bmatrix}^T 
    = \bs{1}_m \otimes \tpop{} + \ui{}{}\\
    \ui{}{} &= \begin{bmatrix}
        \ui{T}{1}
        \ldots
        \ui{T}{m}
    \end{bmatrix}^T\\
    \bs{\epsilon}_i &=
    \begin{bmatrix}
        \epsilon_{i, 1} 
        \ldots
        \epsilon_{i, t}
    \end{bmatrix}^T\\
    \bs{\epsilon} &=
    \begin{bmatrix}
        \bs{\epsilon^T_{1}} 
        \ldots
        \bs{\epsilon^T_{m}}
    \end{bmatrix}^T\\
    \bs{u}_i &\sim \mathcal{N}(\bs{0}, \bs{\Sigma}_{u})\\
    \bs{\epsilon} &\sim \mathcal{N}(\bs{0}, \sigma_{\epsilon}^2\bs{I}_{mt})
\end{align}
}

As specified before, we assume a gaussian prior on the population level term $\tpop{} \sim \mathcal{N}(\muprior, \Sigprior)$. 
The hyper-parameters of the above model, given the definition above, are the noise variance $\sige{2}{}$ and the random effects variance $\Sig{}{u}$. 
Now, at a given decision point $t$, using  estimated values of the hyper-parameters   ($\sigma^2_{\epsilon, t}$ is the estimate of $\sige{2}{}$ and $\Sig{}{u, t}$ is the estimate of $\Sig{}{u}$), the posterior mean and covariance matrix of the parameter $\tparam{}{}$ can be calculated as:
\allowdisplaybreaks
\begin{align}
    \bs{\mu^{(t)}_{\text{post}}} &= \big(\bs{\Tilde{\Sigma}^{-1}_{\theta, t}} + \siget{-2}\bs{A} \big)^{-1} \big( \bs{\Tilde{\Sigma}^{-1}_{\theta, t}} \bs{\mu_{\theta}} + \siget{-2}\bs{B} \big) \label{eqn:postMeanTheta} \\
    \Sig{(t)}{\text{post}} &= \big(\bs{\Tilde{\Sigma}^{-1}_{\theta, t}} + \siget{-2} \bs{A} \big)^{-1} \label{eqn:postCovTheta}
\end{align}
where
\begin{align}
    \bs{A} &= \blockdiag \big( \bs{A_1}, \ldots, \bs{A_m} \big) \\
    \bs{A_i} & = \sum\nolimits_{\tau=1}^t \Phii{}{i\tau} \Phii{T}{i\tau}\\
    \bs{B^T} &= [\bs{B^T_1} \; \ldots \; \bs{B^T_m}]\\
    \bs{B_i} &= \sum\nolimits_{\tau=1}^t \Phii{}{i\tau} R^{(\tau)}_{i}\\
    \bs{\mu^T_{\theta}} &= [\bs{\muprior}^T \; \ldots \; \bs{\muprior}^T ] \label{eqn:mu_t0}\\
    \bs{\Tilde{\Sigma}_{\theta, t}} &= 
    \mathbb{I}_m \otimes \bs{\Sigma_{u,t}} + \mathbb{J}_m \otimes \bs{\Sigprior} \label{eqn:sig_tt}
\end{align}

The action-selection procedure (described in Section \ref{sec:act_select}) uses the Gaussian posterior distribution defined by the posterior mean $\bs{\mu^{(t)}_{\text{post}}}$ and variance $\Sig{(t)}{\text{post}}$ to determine the action selection probability $\pii{t+1}{}$ and the corresponding actions for the next time steps. \\

\noindent \textbf{Hyper-parameter update}: The hyper-parameters in the algorithm's reward model are the noise variance $\sige{2}$ and random effects variance $\Sig{}{u}$. In order to update these variance estimates at the end of decision time $t$, we use Empirical Bayes \cite{morris1983parametric} to maximize the marginal likelihood of observed rewards, marginalized over the parameters $\tparam{}{}$. So, in order to form $\Sig{}{u,t}$ and $\siget{2}$, we solve the following optimization problem:
\begin{align}
    \Sig{}{u,t}, \siget{2} &= \argmax l(\Sig{}{u,t}, \siget{2} ; \mathcal{H}_{1:m}^{(t)})
    \label{eqn:argmaxLL}\\
    \textrm{s.t.} \quad & \Sig{}{u,t} \succ 0\\
      &\siget{2} \geq 0 
\end{align}
where,
\begin{align}
    &l(\Sig{}{u,t}, \siget{2} ; \mathcal{H}) 
    = \log(\det(\bs{X})) - \log(\det(\bs{X} + y \bs{A})) \nonumber \\
    &  + mt \log(y) - y \sum\nolimits_{\tau\in [t]} \sum\nolimits_{i \in [m]} (R^{(\tau)}_{i})^2 - \bs{\mu_{\theta}^T} \bs{X} \bs{\mu_{\theta}} \nonumber \\
    & + (\bs{X} \bs{\mu_{\theta}}  + y \bs{B})^T (\bs{X} + y \bs{A} )^{-1} (\bs{X} \bs{\mu_{\theta}}  + y \bs{B})
\end{align}
Note that, $\bs{X} = \bs{\Tilde{\Sigma}^{-1}_{\theta, t}}$ (see Eq. \ref{eqn:sig_tt}) and $y = \frac{1}{\sigma_{\epsilon, t}^2}$. We solve the optimization problem using gradient descent.

\subsection{Action selection procedure}
\label{sec:act_select}

The action selection procedure utilizes a modified posterior sampling algorithm called the smooth posterior sampling algorithm. Recall from Section~\ref{sec:reward_model}, our model for the reward is a Bayesian linear mixed model with action centering (refer Eq. \ref{eqn:reward_act_center}) where $\pii{t}{i}$ is the probability that the RL algorithm selects action $\action{t}{i} = 1$ in state $\state{t}{i}$ for participant $i$ at decision point $t$. The RL algorithm computes the probability $\pi_i^{(t)}$ as follows:
\begin{equation}
    \begin{aligned}
    \pi_i^{(t)} = \mathbb{E}_{{\Tilde{\beta} \sim \mathcal{N}(\mu_{\text{post}, i}^{(t-1)}, \Sigma_{\text{post}, i}^{(t-1)})}}[\rho(f(\state{t}{i})^T \boldsymbol{\Tilde{\beta}}) |\mathcal{H}_{1:m}^{(t)},  \state{t}{i}]
    \end{aligned}
    \label{eqn:act_select}
\end{equation}
Notice that the last expectation above is  over the draw of $\beta$ from the posterior distribution parameterized by $\boldsymbol{\mu_{\text{post}, i}^{(t-1)}}$ and $\boldsymbol{\Sigma_{\text{post}, i}^{(t-1)}}$ (see Eq. \ref{eqn:postMeanTheta} and Eq. \ref{eqn:postCovTheta} for their definitions).

Classical posterior sampling sets $\rho(x) = \II(x > 0)$. In this case, the posterior sampling algorithm sets randomization probabilities to the posterior probability that the treatment effect is positive. However, when using a pooled algorithm, \citeauthor{zhang2022statistical} showed that between study statistical inference is enhanced if $\rho$ is a \emph{smooth} i.e. continuously differentiable function. 
Using a smooth function ensures that the randomization probabilities formed by the algorithm concentrate. Concentration enhances the replicability of the randomization probabilities if the study is repeated. 
Without concentration, the randomization probabilities might fluctuate greatly between repetitions of the  study \cite{deshpande2018accurate,NEURIPS2021_49ef08ad,zhang2022statistical}. In \trialname{}, we choose $\rho$ to be a generalized logistic function, defined as follows:
\begin{equation}
    \rho(x) = L_{\min} + \frac{ L_{\max} - L_{\min} }{  1 + c \exp(-b x) }
    \label{eqn:smooth_post_sampling0}
\end{equation}
where $c=5$, and $b=21.053$ (please refer to Appendix \ref{sec:smooth_allocation_function} for more details). We set the lower and upper clipping probabilities as $L_{\min} = 0.2$ and $L_{\max} = 0.8$  (i.e., $0.2\le\pi_i^{(t)}\le 0.8$). The probabilities are clipped to facilitate after-study analysis and off-policy evaluation \cite{zhang2022statistical}.

\subsection{Reward Engineering}
\label{sec:reward_design}
To account for delayed effects in the bandit framework, we engineer the reward for the RL algorithm. Note that this engineered reward is only utilized to update the RL algorithm's parameters and hyper-parameters. We are still interested in maximizing the reward defined in Sec. \ref{sec:rl_framework}, and use it to evaluate the algorithm's performance.

The engineered reward $\hat{R}^{(t)}_{i}$ for user $i$ at decision time $t$ is defined as:
\begin{align}
    \hat{R}^{(t)}_{i} &= \reward{t}{i} - \action{t}{i}\text{cost}(\action{t}{i}) \label{eqn:engg_reward}\\
    \text{cost}(\action{t}{i}) &= \lambda \cdot \sigma_{i, \text{obs}}
\end{align}
where $\sigma_{i, \text{obs}}$ is the standard deviation of the observed rewards for a given user $i$, and $\lambda$ is a tuned non-negative hyper-parameter. Note that the reward is not penalized when $\action{t}{i} = 0$. Intuitively, the cost function is designed to allow the RL algorithm to optimize for user engagement, while simultaneously accounting for the delayed effect of sending an intervention message, i.e. $\action{t}{i} = 1$.

{\small
\begin{algorithm}[t]
    \caption{\algorithmname{}}
    \label{algo:multistate_mixed_algo}
    \SetKwInOut{Input}{Input}
    \SetKwInOut{Output}{Output}
    \SetKwData{n}{$D$}
    \SetKwData{d}{$d$}
    \SetKwData{t}{$\tau$}
    \SetKwData{i}{$i$}
    \SetKwData{j}{$j$}
    \SetKwData{priormeanMu}{$\bs{\mu_0}$}
    \SetKwData{priorcovMu}{$\bs{\Sigma_0}$}
    \SetKwData{priorcovB}{$\bs{\Sigma_{u, 0}}$}
    \SetKwData{priorcovE}{$\sigma^2_{\epsilon, 0}$}
    \SetKwData{priormeanTh}{$\bs{\mu_{\theta}}$}
    \SetKwData{priorcovTh}{$\bs{\Sigma_{\theta}}$}
    \SetKwData{OldmeanThti}{$\bs{\mu^{(\d-1)}_{\text{post}, i}}$}
    \SetKwData{OldcovThti}{$\bs{\Sigma^{(\d-1)}_{\text{post}, i}}$}
    \SetKwData{OldmeanTht}{$\bs{\mu^{(\d-1)}_{\text{post}}}$}
    \SetKwData{OldcovTht}{$\bs{\Sigma^{(\d-1)}_{\text{post}}}$}
    \SetKwData{OldcovBt}{$\bs{\Sigma_{u, \d-1}}$}
    \SetKwData{OldcovEt}{$\sigma^2_{\epsilon, \d-1}$}
    \SetKwData{meanTht}{$\bs{\mu^{(\d)}_{\text{post}}}$}
    \SetKwData{covTht}{$\bs{\Sigma^{(\d)}_{\text{post}}}$}
    \SetKwData{covBt}{$\bs{\Sigma_{u, \t}}$}
    \SetKwData{covEt}{$\sigma^2_{\epsilon, \t}$}
    \SetKwData{Actit}{$\pi^{(\t)}_{i}$}
    \SetKwData{Rho}{$\rho$}
    \Input{$m$, \n, $\bs{\mu^{(0)}_{\text{post}}} = \muprior$, $\bs{\Sigma^{(0)}_{\text{post}}} = \Sigprior$, \priorcovB, \priorcovE, $\Rho(x)$}
    \For{\d $=1$ \KwTo $\n $}
    {
        \For{\j $=0$ \KwTo $1$}
        {
            Compute timestep $\t = ((d-1) \times 2) + j$ \\
            \For{\i $=1$ \KwTo $m$}
            {
                Observe state $\state{\t}{i}$\;
                Get posteriors $\OldmeanThti$ and $\OldcovThti$ for user \i from \OldmeanTht and \OldcovTht \;
                Compute action selection probability \Actit using Eq. \ref{eqn:act_select}\\
                Sample action $\action{t}{i} = {\rm Bern} (\Actit)$\\
                Collect reward $R^{(\t)}_{i}$ 
            }
        }
        Update $\Sig{}{u, \d}$ and $\sigma^2_{\epsilon, \d}$ using Eq. \ref{eqn:argmaxLL}, with engineered rewards from Eq. \ref{eqn:engg_reward}\;
        Update posteriors \meanTht and \covTht using Eq. \ref{eqn:postMeanTheta} and Eq. \ref{eqn:postCovTheta}, with engineered rewards from Eq. \ref{eqn:engg_reward}
    }
\end{algorithm}}

\section{Experimental Results}
\label{sec:simulated_exp}

In this section, we detail the design of a simulation testbed (Sec. \ref{sec:simulation_testbed}) to help evaluate the performance of our algorithm. Our experimental setup and the corresponding results are discussed in Sec \ref{sec:exp_results}.

\begin{table*}[ht]
\resizebox{\textwidth}{!}{%
\begin{tabular}{|c|ccccc|ccccc|}
\hline
  &
  \multicolumn{5}{c|}{Minimal Treatment Effect} &
  \multicolumn{5}{c|}{Low Treatment Effect} \\ \cline{2-11} 
 &
  \multicolumn{1}{c|}{} &
  \multicolumn{2}{c|}{HB=Low} &
  \multicolumn{2}{c|}{HB=High} &
  \multicolumn{1}{c|}{} &
  \multicolumn{2}{c|}{HB=Low} &
  \multicolumn{2}{c|}{HB=High} \\ \cline{3-6} \cline{8-11} 
\multirow{-3}{*}{Alg.} &
  \multicolumn{1}{c|}{\multirow{-2}{*}{HB=None}} &
  \multicolumn{1}{c|}{P=50\%} &
  \multicolumn{1}{c|}{P=100\%} &
  \multicolumn{1}{c|}{P=50\%} &
  P=100\% &
  \multicolumn{1}{c|}{\multirow{-2}{*}{HB=None}} &
  \multicolumn{1}{c|}{P=50\%} &
  \multicolumn{1}{c|}{P=100\%} &
  \multicolumn{1}{c|}{P=50\%} &
  P=100\% \\ \hline
\algorithmname &
  \multicolumn{1}{c|}{\cellcolor[HTML]{9AFF99}128.54$\pm$0.18} &
  \multicolumn{1}{c|}{\cellcolor[HTML]{9AFF99}127.23$\pm$0.18} &
  \multicolumn{1}{c|}{\cellcolor[HTML]{FFFFC7}126.01$\pm$0.18} &
  \multicolumn{1}{c|}{\cellcolor[HTML]{FFFFC7}123.22$\pm$0.19} &
  \cellcolor[HTML]{96FFFB}119.55$\pm$0.20 &
  \multicolumn{1}{c|}{\cellcolor[HTML]{9AFF99}129.44$\pm$0.17} &
  \multicolumn{1}{c|}{\cellcolor[HTML]{FFFFC7}128.11$\pm$0.17} &
  \multicolumn{1}{c|}{\cellcolor[HTML]{FFFFC7}126.80$\pm$0.18} &
  \multicolumn{1}{c|}{\cellcolor[HTML]{96FFFB}123.74$\pm$0.19} &
  \cellcolor[HTML]{96FFFB}120.12$\pm$0.20 \\ \hline
BLR &
  \multicolumn{1}{c|}{\cellcolor[HTML]{9AFF99}127.78$\pm$0.16} &
  \multicolumn{1}{c|}{\cellcolor[HTML]{9AFF99}126.60$\pm$0.18} &
  \multicolumn{1}{c|}{\cellcolor[HTML]{FFFFC7}125.78$\pm$0.18} &
  \multicolumn{1}{c|}{\cellcolor[HTML]{FFFFC7}123.23$\pm$0.19} &
  \cellcolor[HTML]{96FFFB}119.60$\pm$0.20 &
  \multicolumn{1}{c|}{\cellcolor[HTML]{9AFF99}129.10$\pm$0.17} &
  \multicolumn{1}{c|}{\cellcolor[HTML]{FFFFC7}127.85$\pm$0.17} &
  \multicolumn{1}{c|}{\cellcolor[HTML]{FFFFC7}126.53$\pm$0.18} &
  \multicolumn{1}{c|}{\cellcolor[HTML]{96FFFB}123.75$\pm$0.19} &
  \cellcolor[HTML]{96FFFB}120.16$\pm$0.20 \\ \hline
random &
  \multicolumn{1}{c|}{\cellcolor[HTML]{9AFF99}127.83$\pm$0.16} &
  \multicolumn{1}{c|}{\cellcolor[HTML]{9AFF99}126.52$\pm$0.18} &
  \multicolumn{1}{c|}{\cellcolor[HTML]{FFFFC7}125.22$\pm$0.18} &
  \multicolumn{1}{c|}{\cellcolor[HTML]{FFFFC7}119.03$\pm$0.21} &
  \cellcolor[HTML]{96FFFB}110.29$\pm$0.23 &
  \multicolumn{1}{c|}{\cellcolor[HTML]{9AFF99}128.97$\pm$0.17} &
  \multicolumn{1}{c|}{\cellcolor[HTML]{FFFFC7}127.70$\pm$0.17} &
  \multicolumn{1}{c|}{\cellcolor[HTML]{FFFFC7}126.45$\pm$0.18} &
  \multicolumn{1}{c|}{\cellcolor[HTML]{96FFFB}120.49$\pm$0.20} &
  \cellcolor[HTML]{96FFFB}112.06$\pm$0.22 \\ \hline
\end{tabular}%
}
\caption{Average total reward per user per simulated trial, averaged across $500$ simulated trials and $120$ users per trial, along with their 95\% confidence intervals (CI) for minimal and low treatment effect settings. HB refers to the level of habituation in the environment, while P is used to denote the proportion of the population who can experience habituation.}
\label{tab:results_min_low}
\end{table*}

\begin{table}[ht]
\resizebox{\columnwidth}{!}{%
\begin{tabular}{|c|c|lc|cc|}
\hline
 &
   &
  \multicolumn{2}{c|}{HB=Low} &
  \multicolumn{2}{c|}{HB=High} \\ \cline{3-6} 
\multirow{-2}{*}{Alg.} &
  \multirow{-2}{*}{HB=None} &
  \multicolumn{1}{c|}{P=50\%} &
  P=100\% &
  \multicolumn{1}{c|}{P=50\%} &
  P=100\% \\ \hline
\algorithmname &
  \cellcolor[HTML]{FFFFC7}132.25$\pm$0.16 &
  \multicolumn{1}{l|}{\cellcolor[HTML]{FFFFC7}130.95$\pm$0.17} &
  \cellcolor[HTML]{FFFFC7}129.71$\pm$0.17 &
  \multicolumn{1}{c|}{\cellcolor[HTML]{96FFFB}124.70$\pm$0.19} &
  \cellcolor[HTML]{96FFFB}121.19$\pm$0.20 \\ \hline
BLR &
  \cellcolor[HTML]{FFFFC7}132.21$\pm$0.16 &
  \multicolumn{1}{l|}{\cellcolor[HTML]{FFFFC7}130.94$\pm$0.17} &
  \cellcolor[HTML]{FFFFC7}129.63$\pm$0.17 &
  \multicolumn{1}{c|}{\cellcolor[HTML]{96FFFB}124.70$\pm$0.19} &
  \cellcolor[HTML]{96FFFB}121.22$\pm$0.19 \\ \hline
random &
  \cellcolor[HTML]{FFFFC7}131.05$\pm$0.17 &
  \multicolumn{1}{l|}{\cellcolor[HTML]{FFFFC7}129.88$\pm$0.17} &
  \multicolumn{1}{l|}{\cellcolor[HTML]{FFFFC7}128.71$\pm$0.17} &
  \multicolumn{1}{c|}{\cellcolor[HTML]{96FFFB}123.19$\pm$0.20} &
  \cellcolor[HTML]{96FFFB}115.39$\pm$0.22 \\ \hline
\end{tabular}%
}
\caption{Average total reward per user per simulated trial along with their 95\% CIs for the high treatment effect settings.}
\label{tab:results_high}
\end{table}

\subsection{Simulation Testbed Design}
\label{sec:simulation_testbed}
We leverage data from the SARA \cite{rabbi2018toward} study, which trialed an mHealth app aimed at sustaining engagement of substance use data collection from participants. Since the SARA study focused on a similar demographic of EAs as the \trialname pilot study, it appears ideal for constructing a simulation testbed. However, note that this data is impoverished. SARA had only $1$ decision point per day, as compared to $2$ per day in \trialname. The goal of the messages sent to the participants in SARA was to increase survey completion in order to collect substance use data. In contrast, the goal of sending intervention messages in \trialname pilot study is to reduce the participant's cannabis use through self-monitoring and mobile health engagement. Moreover, the daily cannabis use data in SARA was collected retro-actively at the end of each week, which often resulted in participant's noisy recollection of their cannabis use, and had missing cannabis use data if the participant chose to not respond. In contrast, participants in \trialname are asked to self-report twice daily, which reduces the amount of missing data if they fail to self-report once. More details and differences are highlighted in Appendix \ref{sec:sara_miwaves}. We construct a \emph{base dataset} of $42$ users after cleaning and imputing the SARA data (please refer to appendix \ref{sec:testbed} for more details).

\noindent\bo{Base Model}: For the base model of the environment, we fit Multinomial Logistic Regression (MLR) models on each of the $42$ users in the base dataset. The learnt weights include weights for the baseline (when action is $0$), and the advantage (added to the baseline when action is $1$). These user models are overfit to learn the user behavior as well capture the noise in the environment. We choose MLR for our user models, as it is interpretable, and performs similar in comparison to a generic neural network (see Appendix \ref{sec:training_user_models}). 

\noindent\bo{Varying Treatment Effects (TE)}: The effect of the intervention message on a particular user is measured by their unique treatment effect size. Given that the intervention messages in SARA had minimal treatment effect \cite{nahum2021translating}, we introduce higher levels of treatment effect into the user models by augmenting their weights. Higher levels of treatment effect increase the likelihood of obtaining higher rewards when taking action $1$. To that end, we construct TE = \emph{low} and TE = \emph{high} treatment effect variants for each MLR user model.
Further details on imputing these effect sizes can be found in Appendix \ref{sec:env_variants}. 

\noindent\bo{Modeling Habituation (HB)}: To account for delayed effects in the environment, we introduce user habituation to repeated stimuli (multiple intervention messages sent to the user in a short span of time) by adding a negative effect in the baseline weights of the MLR user models. To that end, we define \emph{dosage} for each user at each decision point as the weighted average of the number of intervention messages received in the previous six decision points. The weights are decreased with each past decision point, reflecting a diminishing impact of older intervention messages received by the user.
Next, we impute baseline weights for dosage in the MLR user models in a way that higher dosage (more messages received) leads to higher likelihood of generating low rewards, and vice-versa. Note that this procedure simulates how the user may experience habituation; if the RL algorithm does not send many interventions to a user experiencing habituation, the user may dis-habituate and recover their baseline behavior. We construct two environment variants - HB = \emph{low} and HB = \emph{high} habituation effect - by varying the baseline weights for dosage. Additionally, we simulate the proportion of users who can experience habituation within a population - set at either $P=50\%$ or $P=100\%$. Further details on modeling habituation into simulation user models can be found in Appendix \ref{sec:env_variants}.



\subsection{Simulation Results}
\label{sec:exp_results}

\noindent We construct $15$ simulation environment variants using a combination of techniques described in Sec. \ref{sec:simulation_testbed}. For each environment, we simulate $500$ studies with $m=120$ users each, over a period of $D=30$ days ($T=60$). The $m=120$ users are drawn with replacement from the $42$ MLR user models learnt using SARA data.

We compare the performance of our algorithm to two common approaches in mobile health studies. First, is a full pooling algorithm called BLR. BLR utilizes Bayesian Linear Regression \cite{peng2019} to pool data and learn a single model across all the users in a study, and select actions according to the action selection procedure mentioned in Sec. \ref{sec:act_select}. We use engineered rewards (Sec. \ref{sec:reward_design}) to update BLR's parameters and hyper-parameters. We also update BLR hyper-parameters using Empirical Bayes, similar to the approach described in Sec. \ref{sec:online_rl_alg}, for a fair comparison. For both \algorithmname{} and BLR, we update the posteriors at the end of each simulated day (every $2$ decision points), and the hyper-parameters at the end of each week (every $14$ decision points). We refer the reader to Appendix \ref{sec:BLR} for more details about BLR's implementation. In addition to BLR, we also compare against the random algorithm, which utilizes an action selection probability of $\pii{t}{i} = 0.5$.

For each algorithm and simulation environment pair, we calculate the average total reward per user per simulated trial, averaged across the $500$ simulated trials and $120$ users in each trial. We also compute their 95\% confidence intervals (CIs). We summarize our findings in Table \ref{tab:results_min_low} and \ref{tab:results_high}. {In all the simulation environments, reBandit performs no worse than the baseline algorithms.} We highlight the environments where \algorithmname{} significantly outperforms other algorithms (CIs do not overlap) in \emph{green}. In the environments where the CIs for the average total reward overlap for \algorithmname{} and BLR, we individually compare each of the 500 seeded simulations, and count the number of times \algorithmname{} achieved an average total reward per user as compared to BLR. If this number is greater than $50\%$ of the simulations, i.e. greater than $250$, we highlight those environments in \emph{yellow}, otherwise they are highlighted in \emph{blue}. 

The primary takeaway from our simulation results in Tables \ref{tab:results_min_low} and \ref{tab:results_high} is that \algorithmname{} is impactful - it performs better than BLR in most environments, and even in the blue highlighted environments where it performs slightly worse than BLR, the performance is still comparable. It is important to note that our procedures to artificially introduce treatment effects and user habituation into the user models reduces the heterogeneity among the user models. This is due to the fact that our procedure to artificially inject treatment effect establishes a non-negative effect of taking an action across all the users in the user models. The same applies to the procedure for introducing user habituation, as it establishes a clear negative effect with respect to dosage across all users in the user models. However, in practice, higher levels of treatment effect or user habituation effect may lead to more heterogeneity in the population. Given that limitation, it is easy to observe that in our simulations, as levels of treatment effect or user habituation effect are increased, the performance gap between \algorithmname{} and BLR decreases. In simulation environments characterized by more pronounced heterogeneity due to lower levels of treatment and/or habituation effects, \algorithmname{} excels by adeptly identifying and leveraging the heterogeneity within the user population to personalize the likelihood of intervention message delivery and accrues greater rewards.

\vspace{-0.35cm}
\section{Conclusion}
\label{sec:conclusion}
In this paper, we introduced \algorithmname{}, an online RL algorithm which will be a part of the upcoming mobile health study named \trialname aimed at reducing cannabis use among emerging adults.
We addressed the unique challenges inherent in mobile health studies, including limited data, and requirement for algorithmic autonomy and stability, while designing \algorithmname{}. 
We showed that \algorithmname{} utilizes \emph{random-effects} and \emph{informative Bayesian priors} to learn quickly and efficiently in noisy environments which are common in mobile health studies. The introduction of random effects allows \algorithmname{} to leverage the heterogeneity in the study population and deliver personalized interventions. To benchmark our algorithm, we detailed the design of a simulation testbed using prior data, and showed that \algorithmname{} performs equally well or better than two common approaches used in mobile health studies. In the future, we aim to analyze the effectiveness of the interventions in the \trialname{} pilot study. In addition, we aim to investigate the contribution of an individual's data and the study population data towards learning the individual's parameters in the random effects model (see Appendix \ref{sec:induced_states}).

\appendix




{\small
\bibliographystyle{named}
\bibliography{supp}}


\onecolumn
\appendix
\textbf{{\huge Appendices}}

\section{Simulation Testbed}
\label{sec:testbed}
This section details how we transform prior data to construct a dataset, and utilize the dataset to develop the \trialname{} simulation testbed. The testbed is used to develop and evaluate the design of the RL algorithm for the \trialname pilot study. The \emph{base simulator} or the \emph{vanilla testbed} is constructed using the SARA \cite{rabbi2018toward} study dataset. The SARA dataset consists of $N=70$ users, and the SARA study was for 30 days, 1 decision point per day. For each user, the dataset contains their daily and weekly survey responses about substance use, along with daily app interactivity logs and notification logs. We will now detail the procedure to 
construct this base simulator.

\subsection{SARA vs \trialname}
\label{sec:sara_miwaves}
The Substance Use Research Assistant (SARA) \cite{rabbi2018toward} study trialed an mHealth app aimed at sustaining engagement of substance use data collection from participants. Since the SARA study focused on a similar demographic of emerging adults (ages 18-25) as the \trialname pilot study, we utilized the data gathered from the SARA study to construct the simulation testbed. We highlight the key differences between the SARA study and the \trialname pilot study in Table \ref{tab:sara_miwaves}.

\begin{table}[h]
    \centering
    \begin{tabular}{|c|c|}
        \hline
        \bo{SARA} & \bo{\trialname{}}\\
        \hline
        \hline
        $m=70$ & $m=120$\\
        \hline
        $T=30$ & $T=60$\\
        \hline
        \makecell{$\pii{t}{i} = 0.5$} & \makecell{$\pii{t}{i}$ determined by RL}\\
        \hline
        \makecell{Cannabis use data \\self-reported weekly} & \makecell{Cannabis use data \\self-reported twice daily}\\
        \hline
        \makecell{Inspirational and reminder \\messages to increase\\ survey completion and \\collect substance use data} & \makecell{Messages to prompt \\cannabis use reduction \\through self-monitoring \\and improve engagement}\\
        \hline
    \end{tabular}
    \caption{SARA vs \trialname{}: Key differences}
    \label{tab:sara_miwaves}
\end{table}

\subsection{Data Extraction}
\label{sec:data_extract}
First, we will detail the steps to extract the relevant data from the SARA dataset:
\begin{enumerate}
    \item  \bo{App Usage:} The SARA dataset has a detailed log of the user's app activity for each day in the trial, since their first login. We calculate the time spent by each user between a normal entry and an \emph{app paused} log entry, until 12 AM midnight, to determine the amount of time (in seconds) spent by the user in the app on a given day. To determine the time spent by the user in the evening, we follow the same procedure but we start from any logged activity after the 4 PM timestamp, till midnight (12 AM).
    \item \bo{Survey Completion:} The SARA dataset contains a CSV file for each user detailing their daily survey completion status (completed or not). We use this binary information directly to construct the survey completion feature.
    \item \bo{Action:} The SARA dataset contains a CSV file for each user detailing whether they got randomized (with 0.5 probability) to receive a notification at 4 PM, and whether the notification was actually pushed and displayed on the user's device. We use this CSV file to derive the latter information, i.e. whether the app showed the notification on the user's device (yes or no).
    \item \bo{Cannabis Use:} Unlike \trialname{}, the SARA trial did not ask users to self-report cannabis use through the daily surveys. However, the users were prompted to respond to a weekly survey at the end of each week (on Sundays). Through these weekly surveys, the users were asked to retroactively report on their cannabis use in the last week, from Monday through Sunday. We use these weekly surveys to retroactively build the reported cannabis-use for each user in the study. The cannabis use was reported in \emph{grams} of cannabis used, taking values of $0$g, $0.25$g, $0.5$g, $1$g, $1.5$g, $2$g and $2.5$g+. Users who were not sure about their use on a particular day, reported their use as ``\emph{Not sure}''. Meanwhile, users who did not respond to the weekly survey were classified to have ``\emph{Unknown}'' use for the entire week. The distribution of reported cannabis use in SARA across all the users, across all 30 days, can be viewed in Figure \ref{fig:cb_use_dist}.

    \begin{figure}[h]
        \includegraphics[scale=0.5]{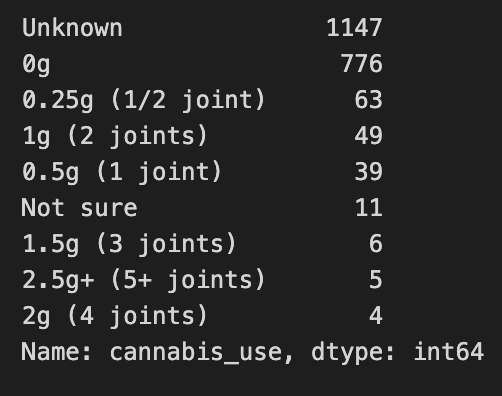}
        \centering
        \caption{Distribution of cannabis use across all users, across all data points}
        \label{fig:cb_use_dist}
    \end{figure}
\end{enumerate}

We build a database for all users using the three features specified above.

\subsection{Data Cleaning}
Next, we will specify the steps to clean this data, and deal with outliers:

\begin{enumerate}
    \item \bo{Users with insufficient data}: We remove users who had more than 20 undetermined (i.e. either ``\emph{Unknown}'' or ``\emph{Not sure}'') cannabis use entries. Upon removing such users, we are left with $N=42$ users. The updated distribution of reported cannabis use in the remaining data across all the users across all 30 days, is demonstrated in Figure \ref{fig:cb_use_dist_filter}.

    \begin{figure}[!h]
        \centering
        \begin{subfigure}[t]{0.49\textwidth}
            \centering
            \includegraphics[width=0.51\textwidth]{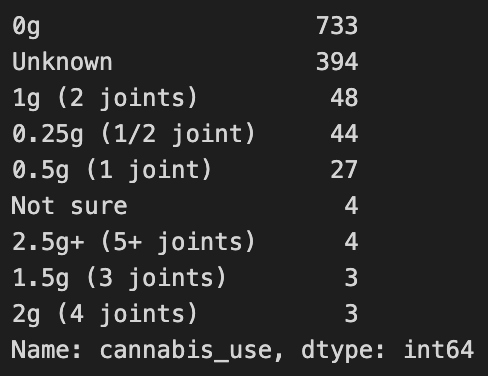}
            \caption{Dataset cannabis-use distribution}
            \label{fig:dataset_cb_dist}
        \end{subfigure}
        \hfill
        \centering
        \begin{subfigure}[t]{0.49\textwidth}
            \centering
            \includegraphics[width=\textwidth]{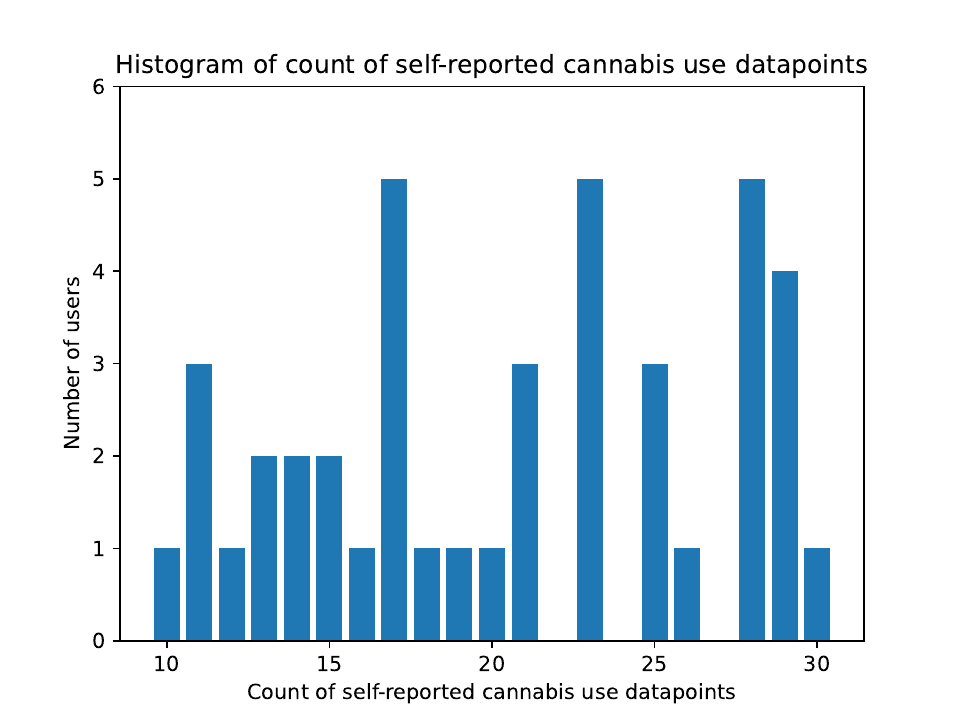}
            \caption{Count of non-missing cannabis-use datapoints vs number of users}
            \label{fig:non-missing_count}
        \end{subfigure}
        \caption{Distribution of cannabis use across all users, across all data points, after removing users with insufficient data}
        \label{fig:cb_use_dist_filter}
    \end{figure}

    \item \bo{Outliers in app usage information}: The average app usage in a given day across all users, comes out to be $244$ seconds (please refer to Section \ref{sec:data_extract} as to how app usage is calculated from the raw data).
    However, due to some technical error during the data collection process, sometimes, a user ended up with greater than $1000$ seconds of app use in a day, with the highest reaching around $67000$ seconds. We observe a similar issue with post-4PM app use data. To deal with such outliers, we clip any post-4PM app use higher than $700$ seconds. There are $20$ such data points in the dataset.
\end{enumerate}

\subsection{Reward computation}
\label{sec:reward_comp}
Next, we calculate the \emph{reward} for each data point in the dataset. It is calculated as follows:
\begin{itemize}
    \item \bo{0}: User did not complete the daily survey, nor used the app.
    \item \bo{1}: User did not complete the daily survey, but used the app outside of the daily survey.
    \item \bo{2}: User completed the survey.
\end{itemize}

We transform the $[0-2]$ reward above to a $[0-3]$ reward defined in \trialname{} (refer Section \ref{sec:rl_framework}) by randomly transforming the data points with $2$ reward to $3$ with $50\%$ probability.

\subsection{Dataset for training user models}
\label{sec:evening_dataset}
We use the post-4PM data (i.e. data from 4 PM to 12 AM midnight) 
(dubbed as the \emph{evening} data) to create a dataset to train the individual user models. The dataset contains the following features:
\begin{itemize}
    \item Day In Study $\in\;[1, 30]$
    \item Cannabis Use $\in\;[0, 2.5g]$
    \item (Evening) App usage $\in\;[0, 700]$
    \item Survey completion (binary)
    \item Weekend indicator (binary)
    \item Action (binary)
    \item Reward $\in \{0, 1, 2, 3\}$
\end{itemize}
We detail the steps to create this \emph{evening} dataset:

\begin{enumerate}
    \item \bo{Evening cannabis-use}: The SARA study documented cannabis use for a given user for a whole day. In contrast, \trialname{} will be asking users to self-report their cannabis use twice a day. To mimic the same, after discussions among the scientific team,
    we split a given user's daily cannabis use from SARA into morning and evening use, and multiply a factor of $0.67$ to generate their evening cannabis use. Also, the \trialname{} study will be recruiting users who typically use cannabis at least 3 times a week. We expect their use to be much higher than that of users in SARA. So, we multiply the evening cannabis use by a factor of $1.5$. Thus, we generate the evening cannabis use from the user's daily use reported in SARA as follows:
    \begin{equation}
        \text{Evening CB Use} = \text{That Day's SARA CB Use} \times 0.67 \times 1.5
    \end{equation}
    
    \item \bo{Feature normalization}: The resulting dataset's features are then normalized as follows:
    \begin{itemize}
        \item \bo{Day in study} is normalized into a range of $[-1, 1]$ as follows
        \begin{align}
            \text{Day in study (normalized)} = \frac{\text{Day in study} - 15.5}{14.5}
        \end{align}
        \item \bo{App usage} is normalized into a range of $[-1, 1]$ as follows
        \begin{align}
            \text{App usage (normalized)} = \frac{\text{App usage} - 350}{350}
        \end{align}
        \item \bo{Cannabis use} (evening) is normalized into a range of $[-1,1]$ as follows
        \begin{align}
            \text{Cannabis use (normalized)} = \frac{\text{Cannabis use} - 1.3}{1.35}
        \end{align}
    \end{itemize}
\end{enumerate}

\subsection{Training User Models}
\label{sec:training_user_models}

As specified above in Sec. \ref{sec:evening_dataset}, we use to \emph{evening} dataset to train our user models. This dataset has the following features:  
\begin{itemize}
    \item Day In Study $\in\;[-1, 1]$: Negative values refer to the first half of the study, while positive values refer to the second half of the study. A value of $0$ means that the user is in the middle of the study. $-1$ means that the user has just begun the study, while $1$ means they are at the end of the 30 day study. 
    \item Cannabis Use $\in\;[-1, 1]$: Negative values refer to the user's cannabis use being lower than the population's average cannabis use value, while positive values refer to user's cannabis use being higher than the study population's average cannabis use value. A value of $0$ means that the user's cannabis use is the average value of cannabis use in the study population. Meanwhile, $-1$ means that the user is not using cannabis, and $1$ means that the user used the highest amount of cannabis reported by the study population. 
    \item (Evening) App usage $\in\;[-1, 1]$: Negative values refer to the user's app use being lower than the population's average app use value, while positive values refer to user's app use being higher than the study population's average app use value. A value of $0$ means that the user's app usage is the average amount of app usage observed in the study population. Meanwhile, $-1$ means that the user's app usage is non-existent (i.e. zero). On the other hand, $1$ means that the user's app usage is the highest among the observed app usage values in the study population. 
    \item Survey completion $\in \{0, 1\}$: A value of $0$ refers to the case where the user has not finished the decision point's EMA, while a value of $1$ refers to the case where the user has responded to the decision point's EMA.
    \item Weekend indicator $\in \{0, 1\}$: A value of $0$ refers to the case where the decision point falls on a weekday, while a value of $1$ refers to the case where the decision point falls on a weekend.
    \item Action $\in \{0, 1\}$: A value of $0$ means that action was not taken (i.e. no notification or intervention message was shown), while $1$ means that an action was taken (i.e. a notification or intervention message was shown).
    \item Reward $\in \{0, 1, 2, 3\}$: Same as defined in Section \ref{sec:reward_comp}
\end{itemize}

We morph these features into a $12$ dimensional feature vector, defined as follows:
\begin{enumerate}
    \item Intercept (1)
    \item Survey Completion
    \item Standardized App Usage
    \item Standardized Cannabis Use
    \item Weekend Indicator
    \item Standardized Day in Study
    \item Action taken multiplied by the Intercept (1)
    \item Action taken multiplied by Survey Completion
    \item Action taken multiplied by Standardized App Usage
    \item Action taken multiplied by Standardized Cannabis Use
    \item Action taken multiplied by Weekend Indicator
    \item Action taken multiplied by Standardized Day in Study
\end{enumerate}

In the rest of the section, the weights corresponding to features $1-6$ are referred to as the baseline weights, and the weights for $7-12$ are referred to as the advantage weights.

We fit the reward using our user models. Before we train user models, we do a \emph{complete-case analysis} on the \emph{evening} dataset. It involves removing all the data points which have \bo{any} missing feature. This can either be a missing ``cannabis use'' value, or a missing ``action'' (which is the case on the first day of the study). 
\\\\
Given that our target variable i.e. the reward is categorical (0-3) in nature, we consider two options for our generative user models:
\begin{itemize}
    \item \bo{Multinomial Logistic Regression (MLR)} - We fit a multinomial logistic regression model on our dataset. Given $K$ classes (K=4 in our case), the probability for a data point $i$ belonging to a class $c$ is given by:

    \begin{align}
        P(Y_i = c|X_i) = \frac{e^{\beta_c \cdot X_i}}{\sum_{j=1}^{K} e^{\beta_j \cdot X_i}}
    \end{align}
    where $X_i$ are the features of the given \emph{data point} $i$, and $\beta_j$ is the learned coefficient for a given reward class $j$. Each \emph{data point} refers to a user-decision point, we use these terms interchangibly throughout the document. 
    
    The model is optimized using the following objective:
    \begin{align}
        \min_{\beta} - \sum_{i=1}^{n} \sum_{c=1}^{K} \mathbbm{1}_{\{Y_i = c\}} \log (P(Y_i = c|X_i)) + \frac{1}{2} ||\beta||^2_{F}
    \end{align}
    
     We use python's \texttt{scikit-learn} package for training the multinomial logistic regression model (more information \href{https://scikit-learn.org/stable/modules/linear_model.html#multinomial-case}{here}). It makes sure that $\sum_{j} \beta_j = 0$. We use the following parameters for training the model using \texttt{scikit-learn}: 
    \begin{itemize}
        \item \bo{Penalty:} L2
        \item \bo{Solver:} LBFGS
        \item \bo{Max. iterations:} 200
        \item \bo{Multi-class:} Multionomial
    \end{itemize}
    \item \bo{Multi-layer perceptron (MLP) Classifier} - We fit a simple neural network on our dataset, and use the last layer's logits as probabilities for each class.
    
    We use python's \texttt{scikit-learn} package for training the MLP Classifier. We use the following parameters for training the model - 
    \begin{itemize}
        \item \bo{Hidden layer configuration:} (7, )
        \item \bo{Activation function:} Logistic
        \item \bo{Solver:} Adam
        \item \bo{Max. iterations:} 500
    \end{itemize}
\end{itemize}

\begin{figure}[t]
    \centering
    \includegraphics[width=0.99\textwidth]{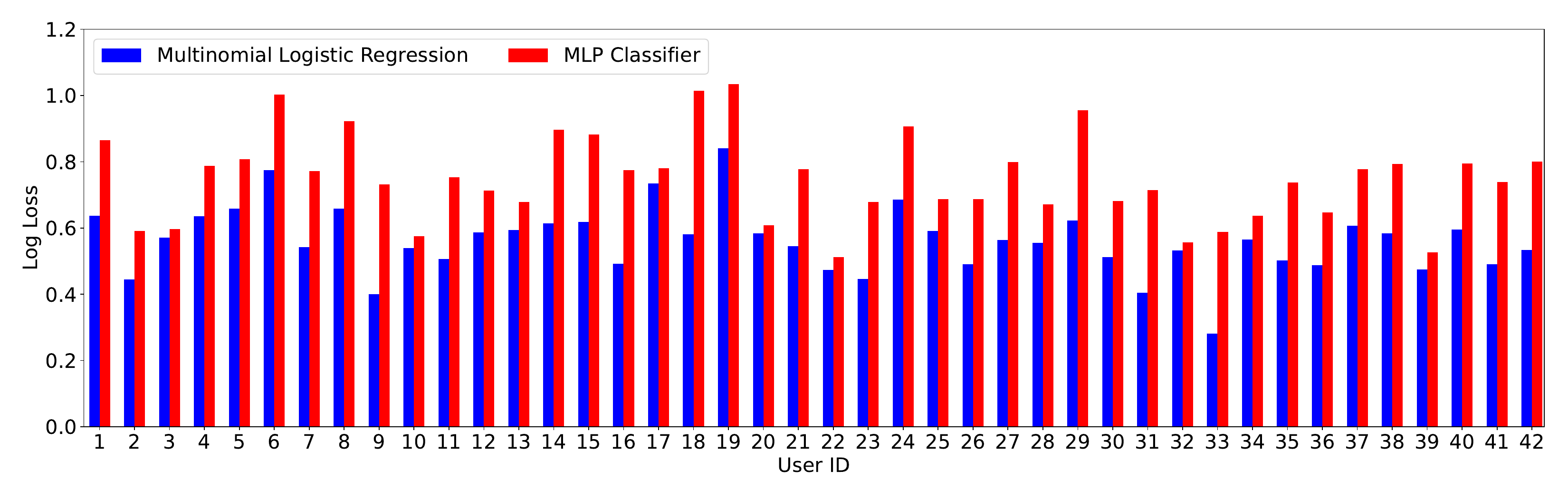}
    \caption{Comparison of log loss between the two models across all users}
    \label{fig:log_loss}
\end{figure}

We choose MLR for our user models, as it is interpretable, and offers similar (if not better) performance as compared to a generic neural network (see Figure \ref{fig:log_loss}). Interpretability is important here since we would like to vary the treatment effect in our generative user models.

\begin{figure}[ht]
    \centering
    \includegraphics[width=\textwidth]{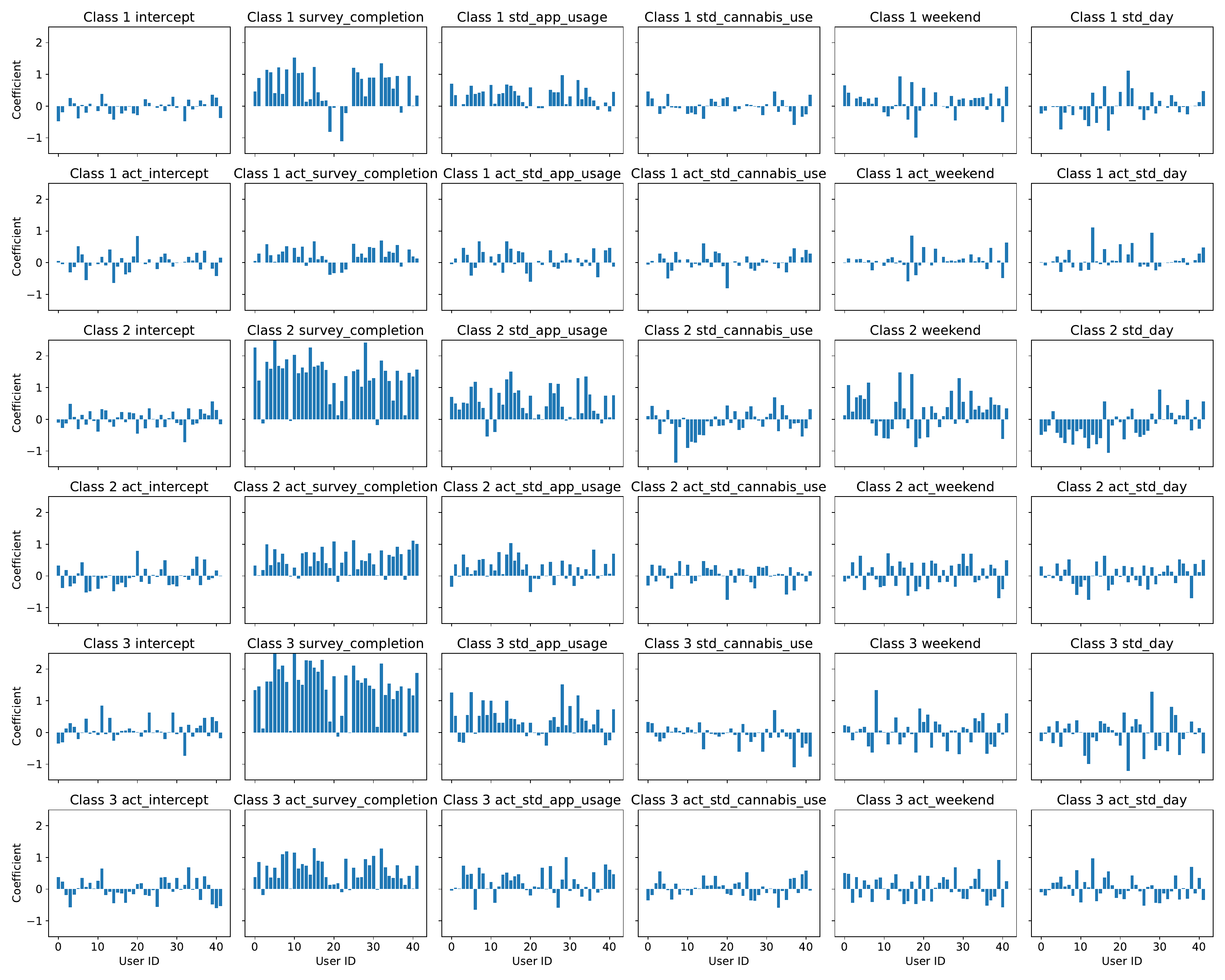}
    \caption{\textbf{Bar plot of coefficients of features in the MLR user models relative to coefficients of class 0, across all $N=42$ users.}}
    \label{fig:coeff_relative}
\end{figure}

Figure \ref{fig:coeff_relative} represents the learnt coefficients of the MLR user model for classes 1 to 3, relative to class 0's coefficients. Note that both \emph{survey completion} and \emph{app usage} seem to exhibit strong relationship wrt the target variable for most of the users. To be specific, in Figure \ref{fig:coeff_relative}, coefficients of both \emph{survey completion} and \emph{app usage} are mostly positive across most of the $N=42$ users, both in baseline and advantage. The magnitude of the relative weights of these features keeps increasing as the reward class increases. This signifies that if a user is engaged (completing surveys, using the app), they are more likely to generate a non-zero reward as compared to a reward of 0. We use the probabilities as weights from the MLR models to stochastically generate the reward during the data generation process. More on that in Section \ref{sec:data_gen}.

\subsection{Dataset for generative process}
Next, we will create a dataset for the generative process for a simulation. To that end, we impute missing values in the \emph{evening} dataset, and also create the \emph{morning} dataset. We describe the procedure for both below.

\begin{itemize}
    \item \bo{Imputation in \emph{evening} dataset}: As stated before, the vanilla \emph{evening} dataset has a few missing values for ``cannabis use'' and ``action'' values. We impute the values for ``cannabis use'' as follows - for a given missing value, we first determine the day of the week, and then replace the unknown/missing value with the average of the cannabis use across the available data of the user for that day of the week.
    
    Note that during the simulation, the action used by the user models for reward generation will be determined by the RL algorithm in the simulation. Hence, we do not need to impute ``action'' here.

    \item \bo{Generating \emph{morning} dataset}: Similar to the \emph{evening} dataset, we generate a \emph{morning} dataset per user to mimic the data we would receive from \trialname{} (2 decision points a day). We generate the following features:
    \begin{itemize}
        \item \bo{Cannabis Use}: We generate the morning cannabis use as follows:
        
        \begin{equation}
            \text{Morning CB Use} = \text{That Day's SARA CB Use} \times 0.33 \times 1.5
        \end{equation}
        \item \bo{App Usage}: Since there are less than 30 evening app usage values per user in the SARA dataset, we decide to use these values as an empirical distribution, and resample, with replacement, from the 30 values for each user at each morning. The sampled value is the user's  morning app usage.

        \item \bo{Survey Completion}: For each user, we determine the proportion of times they responded to the daily survey during the SARA study. Using this ratio as our probability of survey completion, we sample from a binomial distribution to construct the Survey Completion feature for the morning dataset for each user for each day.

        \item \bo{Day In Study and Weekend indicator}: We create one morning data point per day, so the day in study and weekend indicator features mimic the evening dataset. 
    \end{itemize}
\end{itemize}

\subsection{Simulations: Data generation}
\label{sec:data_gen}
In this section, we detail our data generation process for a simulation. We assume that any given user $i$, her trajectory in an RL simulation environment with $T$ total decision points has the following structure: $\mathcal{H}_{i}^{(T)} = \{\state{1}{i}, \action{1}{i}, \reward{1}{i}, \cdots, \state{T}{i}, \action{T}{i}, \reward{T}{i}\}$.

We also assume that the \emph{combined} dataset has the following structure - for any given decision point $t$ ($t \in [1,60])$, the state, $\state{t}{i}$, is constructed (partially) based on: the user's \emph{app usage} from $t-1$ to $t$, the \emph{survey completion} indicator (whether user fills the survey after receiving action at $t$) for decision point $t$, and the \emph{cannabis use} of the user from $t-1$ to $t$. The data at decision point $t$ from the \emph{combined} dataset is used to generate $\reward{t}{i}$, which in turn helps generate features to form $\state{t+1}{i}$ (refer Section \ref{sec:rl_framework}).

\begin{enumerate}
    \item Given a set number of users (parameter of the simulator) to simulate, we sample users with replacement from the $N=42$ users in the \emph{combined} dataset.
    \item We start the simulation in the morning of a given day. From decision point $t=1$ to $T$ in the simulation ($T=60$), for each sampled user $i$ (refer to the RL framework in Section \ref{sec:rl_framework}):
    \begin{enumerate}
        \item If $t>1$, given previously-generated reward $\reward{t-1}{i}$, we construct the following features - \emph{survey completion}, \emph{app usage indicator}, and \emph{activity question} (which will help form $\state{t}{i}$ according to Section \ref{sec:rl_framework}), according to the following rules:
        \begin{align}
            \text{Survey Completion} = 
            \left\{
        	\begin{array}{ll}
        		1  & \mbox{if } \reward{t-1}{i} \geq 2 \\
        		0 & \mbox{if } \reward{t-1}{i} < 2
        	\end{array}
            \right.
            \label{eqn:survey_comp}
        \end{align}
        \begin{align}
            \text{App Usage Indicator} = 
            \left\{
        	\begin{array}{ll}
        		1  & \mbox{if } \reward{t-1}{i} \geq 1 \\
        		0 & \mbox{if } \reward{t-1}{i} = 0
        	\end{array}
            \right.
            \label{eqn:app_use}
        \end{align}
        \begin{align}
            \text{Activity Question} = 
            \left\{
        	\begin{array}{ll}
        		1  & \mbox{if } \reward{t-1}{i} = 3 \\
        		0 & \mbox{otherwise}
        	\end{array}
            \right.
            \label{eqn:act_ques}
        \end{align}
        
        We communicate the aforementioned calculated features (Equation \ref{eqn:survey_comp}, \ref{eqn:app_use}, and \ref{eqn:act_ques}), along with the \emph{cannabis use} from $t-2$ to $t-1$ to the RL algorithm. This is similar to how a user would self-report and convey this information to the app, to help the algorithm derive the user current state $\state{t}{i}$.

        At $t=1$, since there is no $\reward{0}{i}$, note that the RL algorithm uses some initial state $\state{1}{i}$. Since we intend to simulate \trialname{}, all RL algorithms will use the following $\state{1}{i}$:
        \begin{itemize}
            \item $S_{i, 1}^{(1)}$: Set to 0. At the start of the trial, there is no engagement by the user.
            \item $S_{i, 2}^{(1)}$: Set to 0. The first decision point is the morning, which corresponds to 0.
            \item $S_{i, 3}^{(1)}$: Set to 1. The users in \trialname{} use cannabis regularly (at least 3x a week), so we ascertain the users in the trial to have high cannabis use at the start of the trial.
        \end{itemize}
        
        \item We ask the RL algorithm for the action $\action{t}{i}$ to be taken at the current decision point $t$ for user $i$.
        \item We take the data from the $t^{th}$ decision point in \emph{combined} dataset of the user $i$ - specifically the user's \emph{survey completion} at $t$, the user's \emph{app usage} from $t-1$ to $t$, and the user's \emph{cannabis use} from the $t-1$ to $t$. We calculate the \emph{weekend indicator} by checking whether the decision point $t$ falls on Saturday or Sunday. We feed this data, along with the action $\action{t}{i}$ from the previous step, to the user $i$'s trained user model from Section \ref{sec:training_user_models} to obtain a reward $\reward{t}{i}$.
        
    \end{enumerate}
\end{enumerate}

\subsection{Environment Variants}
\label{sec:env_variants}
We will now describe the design of different environment variants for our simulator, and how we go about operationalizing them. Since we know that the \trialname{} pilot study will have 120 participants, all our simulation environments will have $N=120$ participants. We sample these participants with replacement from the $N=42$ participants from the \emph{combined dataset} during the data generation process.

\begin{enumerate}
    \item \bo{Varying size of the treatment effect:} We construct three variants wrt the size of the treatment effect:
    \begin{itemize}
        \item \bo{Minimal treatment effect size}: We keep the treatment effect i.e. advantage coefficients in the MLR model that we learn from the users in the SARA data.
    \end{itemize}
    
    For the other two variants, we set the advantage intercept weight for each of the MLR user models as follows: we find the minimum advantage intercept weight across all classes learnt using the SARA user data, and if it is not assigned to class 0 - we swap the weight with that of class 0. Then we set the advantage intercept weight of class 2 and class 3 to be the average of both.
    The reason we do so is because that we believe that at any given point, taking an action will always have a non-negative immediate treatment effect. To that end, at any given time, taking an action will always result in higher probabilities for generating a non-zero reward, and lower probabilities for generating a zero reward, as compared to not taking an action. Setting class weights by assigning the minimum weight to reward class 0 helps us achieve that. Also, since we trained the user models from SARA data where we assigned 2 and 3 reward with equal probability wherever reward of 2 was assigned, we set their corresponding advantage intercept weights to be equal. Hence, we set them to be the average of the observed weights of the two reward classes. Moreover, we decide to work with the user's learnt weights from SARA to preserve heterogeneity among users. 

    We then multiply the advantage intercept weights for all the classes by a factor, which we refer to as the \emph{multiplier}. We select the multipliers for our other variants after observing the standardized advantage intercept weights. To do so, for each multiplier, we first generate a dataset of trajectories for each of the $N=42$ users, by sampling action with probability $0.5$. We generate 500 sets of such trajectories (i.e. 500 simulations with $N=42$ users in each simulation). For each set of trajectories (i.e each simulation) of all $N=42$ users, we fit a GEE linear model with fixed effects and robust standard errors. The model is given as:
    \begin{align}
        \notag E\left[R_{i}^{(t+1)}|S_1, S_2, S_3, a\right] &= \alpha_0 + \alpha_1 S_1 + \alpha_2 S_2 + \alpha_3 S_3 + \alpha_4 S_1 S_2 + \alpha_5 S_1 S_3 + \alpha_6 S_2 S_3 + \alpha_7 S_1 S_2 S_3\\
        &+ a  (\beta_0 + \beta_1 S_1 + \beta_2 S_2 + \beta_3 S_3 + \beta_4 S_1 S_2 + \beta_5 S_1 S_3 + \beta_6 S_2 S_3 + \beta_7 S_1 S_2 S_3).
        \label{eqn:sanity_check_reward_model}
    \end{align}
    
    \begin{figure}[!ht]
        \centering
        \includegraphics[width=0.45\textwidth]{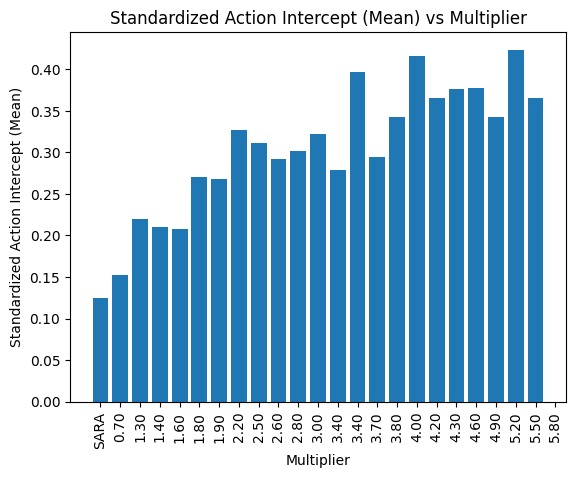}
        \caption{Comparison of mean standardized action intercept weight vs advantage intercept multiplier. Mean is taken across 500 simulations with $N=42$ users (from SARA) in each simulation}
        \label{fig:standardized_act_intercept}
    \end{figure}

    We take the learned weight of $\beta_0$ and divide it by the sample standard deviation of observed rewards for that set of trajectories (i.e. for that simulation) of $N=42$ users - in order to obtain the \emph{standardized action intercept weight}. We do so for all the different multipliers we consider, and the results are summarized in Fig \ref{fig:standardized_act_intercept}. We first check the minimal effect (weights from SARA), and see that the standardized treatment effect for the minimum effect setting comes out to be around $0.12$. We want our low treatment effect setting to be closer to $0.15$ and our higher to be around $0.3$. So we tinker our weights as mentioned in the procedure above, and re-run the simulations. Since we do not scale the weights (yet) by any factor, this is equivalent to a multiplier setting of $1.0$. We observe a standardized treatment effect of $0.18$ in this case. Using this as a reference point, we try a lower and higher multiplier, and we observe that $0.7$ and $2.5$ give us the desired standardized treatment effect for low effect and high effect settings respectively.
    
    Now, the remaining variants are described as follows:
    \begin{itemize}
        \item \bo{Low}:  We multiply the advantage coefficients for each of the MLR user models for each class by $0.7$.
        
        \item \bo{High}: We multiply the advantage coefficients for each of the MLR user models (as mentioned above) for each class by $2.5$. This way, we are increasing the size of all advantage intercepts in the MLR user model, and in turn further skews the probability of getting a higher reward when taking an action as compared to a zero reward.
    \end{itemize}

    \bo{Discussion:} Given the way we model our user models using MLR, they are not complex enough to support higher standardized effect sizes (through advantage intercepts). Recall that the probability of a data point belonging to a particular reward class $c$ is defined using our MLR model as follows:
    
    \begin{align*}
        P(Y_i = c|X_i) = \frac{e^{\beta_c \cdot X_i}}{\sum_{j=1}^{K} e^{\beta_j \cdot X_i}}
    \end{align*}

    Note that we use a linear model in the exponent of the above MLR model, i.e. $\beta_j X_i$ is linear. This is due to our design choices of trying to keep our models simple and interpretable. Since the weights are constrained (always sum up to 1), there is a threshold upto which one can increase the standardized treatment effect through these user models. We see this behavior in Figure \ref{fig:standardized_act_intercept}, where we observe that the standardized effect size hovers around $0.35 - 0.4$, even though we are scaling the advantage intercept weights by larger multipliers.

    \item \bo{Habituation (affecting the baseline effect)}: In order to incorporate user habituation to repeated stimuli (i.e. intervention messages) into the user models, we first define \emph{dosage} for a user $i$ at time $t$ as $Q^{(t)}_{i} = d_{\kappa} \sum_{j=1}^{6} \kappa^{j-1} \action{t-j}{i}$, i.e. weighted average of the number of actions sent to the user $i$ in the recent 6 decision points prior to decision point $t$. We set $\kappa = \frac{5}{6}$ to represent looking back 6 decision points, and scale each sum by a constant $d_{\kappa} = \frac{1-\kappa}{1-\kappa^6}$ so that weights add up to 1. To incorporate habituation into the MLR user models, we add a new baseline weight for dosage. For a given reward class $c$ (where $c \in \{0, 1, 2, 3\}$), we set this weight in the MLR user model as follows:
    \begin{align}
        \beta_{c, \text{dosage}} = 
            \begin{cases} 
                \frac{\sum\limits^{6}_{i = 1} \beta_{c, i}}{\eta} & \text{if} \;\;\; \sum\limits^{6}_{i = 1} \beta_{0, i} \geq 0\\
                -\frac{\sum\limits^{6}_{i = 1} \beta_{c, i}}{\eta} & \text{if} \;\;\; \sum\limits^{6}_{i = 1} \beta_{0, i} < 0\\
            \end{cases}
    \end{align}
    
    where the numerator is the sum of all the baseline weights (since there are 6 baseline weights, see Appendix \ref{sec:training_user_models}), and $\eta$ is used to control the intensity of habituation. Notice that we flip the sign of the weight if the sum of the baseline weights for reward class $0$ is negative. This is done to ensure that higher dosage (i.e. higher levels of repeated stimuli in the near past) leads to an increase in likelihood of generating lower rewards (namely reward classes $0$ and $1$). We set the weight in this way (by summing the baseline weights) to keep the user models as heterogeneous as possible. However, since all the user models now have a clear negative effect wrt dosage, these modified user models are perceived to be less heterogeneous than the ones originally learnt from the SARA data.

    We construct two variants by modifying the value of $\eta$, to vary the intensity of habituation in the user models:
    \begin{itemize}
        \item \bo{Low}: Low intensity habituation effect is set by using $\eta = 6$.
        \item \bo{High}: High intensity habituation effect is set by using $\eta = 1$.
    \end{itemize}

    The higher the habituation effect, the higher the likelihood of getting lower rewards when dosage $Q^{(t)}_{i}$ is non-zero.

    \item \bo{Proportion of Users with habituation}: For the environments with habituation, we vary the proportion of the population who can experience habituation. To that end, we construct two variants with respect to the proportion of the simulated study population who can experience habituation (i.e. proportion of the simulated population whose models are augmented to have weights for dosage as described above) - (i) $50\%$ and (ii) $100\%$



\end{enumerate}

Using a combination of the user model variants mentioned above, we design a total of $15$ environment variants:
\begin{enumerate}
    \item Minimal treatment effect without habituation
    \item Minimal treatment effect with low habituation in 50\% of the population
    \item Minimal treatment effect with low habituation in 100\% of the population
    \item Minimal treatment effect with high habituation in 50\% of the population
    \item Minimal treatment effect with high habituation in 100\% of the population
    \item Low treatment effect without habituation
    \item Low treatment effect with low habituation in 50\% of the population
    \item Low treatment effect with low habituation in 100\% of the population
    \item Low treatment effect with high habituation in 50\% of the population
    \item Low treatment effect with high habituation in 100\% of the population
    \item High treatment effect without habituation
    \item High treatment effect with low habituation in 50\% of the population
    \item High treatment effect with low habituation in 100\% of the population
    \item High treatment effect with high habituation in 50\% of the population
    \item High treatment effect with high habituation in 100\% of the population
\end{enumerate}


\section{Full-pooling algorithm - Bayesian Linear Regression (BLR)}
\label{sec:BLR}
For a given user $i$ at decision point $t$, the action-centered training model for Bayesian Linear Regression (BLR) is defined as:

\begin{align}
    R_i^{(t)} = g(S_i^{(t)})^T \boldsymbol{\alpha}  + (a_i^{(t)} - \pi_i^{(t)}) f(S_i^{(t)})^T \boldsymbol{\beta} + (\pi_i^{(t)})f(S_i^{(t)})^T \boldsymbol{\gamma} + \epsilon_i^{(t)}
\end{align}
where $\pi_i^{(t)}$ is the probability of taking an active action i.e. sending an intervention message ($a_i^{(t)} = 1$). $g(S_i^{(t)})$ and $f(S_i^{(t)})$ are the baseline and advantage vectors respectively. $\epsilon_i^{(t)}$ is the error term, assumed to be independent and identically distributed (i.i.d.) Gaussian noise with mean 0 and variance $\sigma_{\epsilon}^2$.

The probability $\pi_i^{(t)}$ of sending a intervention message ($a_i^{(t)} = 1$) is calculated as:
\begin{align}
    \pi_i^{(t)} = \mathbb{E}_{{\Tilde{\beta} \sim \mathcal{N}(\mu_{\text{post}}^{(t-1)}, \Sigma_{\text{post}}^{(t-1)})}}[\rho(f(S_i^{(t)})^T \boldsymbol{\Tilde{\beta}}) |\mathcal{H}_{1:m}^{(t-1)},  S_{i}^{(t)}]
\end{align}

where $\mathcal{H}_{i}^{(T)} = \{\state{1}{i}, \action{1}{i}, \reward{1}{i}, \cdots, \state{T}{i}, \action{T}{i}, \reward{T}{i}\}$ is the trajectory for a given user $i$ upto time $t$.

The Bayesian model requires the specification of prior values for $\alpha$, $\beta$ and $\gamma$. While we do not assume the terms $\alpha$, $\beta$ and $\gamma$ to be independent, we have no information about the correlation between the $\alpha$, $\beta$ and $\gamma$ terms, and hence we set the correlation to be 0 in our prior variance $\Sigma_{\text{prior}}$. Hence, ${\boldsymbol{\Sigma}}_{\text{prior}} = \text{diag}(\boldsymbol{\Sigma_{\alpha}}, \boldsymbol{\Sigma_{\beta}}, \boldsymbol{\Sigma_{\beta}})$ is the prior variance of all the parameters (note that the posterior var-covariance matrix will have off diagonal elements).  Next ${\boldsymbol{\mu}}_{\text{prior}} = (\boldsymbol{\mu_{\alpha}}, \boldsymbol{\mu_{\beta}}, \boldsymbol{\mu_{\beta}})$ is the prior mean of all the parameters.  The prior of each parameter is assumed to be normal and given as:
\begin{align}
    \boldsymbol{\alpha} \sim \mathcal{N}(\boldsymbol{\mu_{\alpha}}, \boldsymbol{\Sigma_{\alpha}})\\
    \boldsymbol{\beta} \sim \mathcal{N}(\boldsymbol{\mu_{\beta}}, \boldsymbol{\Sigma_{\beta}})\\
    \boldsymbol{\gamma} \sim \mathcal{N}(\boldsymbol{\mu_{\beta}}, \boldsymbol{\Sigma_{\beta}})
\end{align}

Since the priors are Gaussian, and the error term is Gaussian, the posterior distribution of all the parameters, given the history of state-action-reward tuples (trajectory) up until time $t$ is also Gaussian. Let us denote all the parameters using $\boldsymbol{\theta}^T = (\boldsymbol{\alpha}^T, \boldsymbol{\beta}^T, \boldsymbol{\gamma}^T)$. The posterior distribution of $\boldsymbol{\theta}$ given the current history $\mathcal{H}^{(t)}$ = $\{\state{\tau}{}, \action{\tau}{}, \reward{\tau}{}\}_{\tau \leq t}$ (data of all users) is denoted by $\mathcal{N}(\mu_{\text{post}}^{(t)}, \Sigma_{\text{post}}^{(t)})$, where:
\begin{align}
    \boldsymbol{\Sigma}_{\text{post}}^{(t)} &= \bigg( \frac{1}{\sigma_{\epsilon}^2} \boldsymbol{\Phi}_{1:t}^T \boldsymbol{\Phi}_{1:t} + {\boldsymbol{\Sigma}}^{-1}_{\text{prior}} \bigg)^{-1}\\
    \boldsymbol{\mu}_{\text{post}}^{(t)} &= \boldsymbol{\Sigma}_{\text{post}}^{(t)} \bigg( \frac{1}{\sigma_{\epsilon}^2} \boldsymbol{\Phi}_{1:t}^T \mathbf{R}_{1:t} + {\boldsymbol{\Sigma}}^{-1}_{\text{prior}} {\boldsymbol{\mu}}_{\text{prior}} \bigg)
\end{align}

$\boldsymbol{\Phi}_{1:t}$ is a stacked vector of $\Phi{}{}(S, a)^T = [ g(\state{t}{})^T, (\action{t}{} - \pii{t}{}) f(\state{t}{})^T, \pi^{(t)}f(\state{t}{})^T ]$ for all users, for all decision points from 1 to $t$. $\mathbf{R}_{1:t}$ is a stacked vector of rewards for all users, for all decision points from 1 to $t$. $\sigma_{\epsilon}^2$ is the noise variance, ${\boldsymbol{\Sigma}}_{\text{prior}} = \text{diag}(\boldsymbol{\Sigma_{\alpha}}, \boldsymbol{\Sigma_{\beta}}, \boldsymbol{\Sigma_{\beta}})$ is the prior variance of all the parameters, and ${\boldsymbol{\mu}}_{\text{prior}} = (\boldsymbol{\mu_{\alpha}}, \boldsymbol{\mu_{\beta}}, \boldsymbol{\mu_{\beta}})$ is the prior mean of all the parameters.

We initialize the noise variance by $\sigma_{\epsilon}^2=0.85$; see 
Equation (\ref{eqn:multistate_optimization_fixed}) for the update equation. See Appendix \ref{sec:priors} for the choice of $0.85$. Then we
update the noise variance by solving the following optimization problem, that is maximizing the marginal (log) likelihood of the observed rewards, marginalized over the parameters:
\begin{align}
    \sigma_{\epsilon}^2 &= \argmax[- \log(\det(\boldsymbol{X} + y \boldsymbol{A})) + mt \log(y) - y \sum_{\tau\in [t]} \sum_{i\in [m]} (R_{i}^{(\tau)})^2 \nonumber \\
    &+ (\boldsymbol{X} \boldsymbol{\mu_{\text{prior}}}  + y \boldsymbol{B})^T (\boldsymbol{X} + y \boldsymbol{A} )^{-1} (\boldsymbol{X} \boldsymbol{\mu_{\text{prior}}}  + y \boldsymbol{B})]
    \label{eqn:multistate_optimization_fixed}
\end{align}
where $\boldsymbol{X} = \boldsymbol{\Sigma}^{-1}_{\text{prior}}$, $y = \frac{1}{\sigma_{\epsilon}^2}$,  $\boldsymbol{A} = \boldsymbol{\Phi}_{1:t}^T \boldsymbol{\Phi}_{1:t}$, $\boldsymbol{B} = \boldsymbol{\Phi}_{1:t}^T \mathbf{R}_{1:t}$, $t$ is current total number of decision points, and $m$ is the total number of users.

\section{Baseline and Advantage Functions}
\label{sec:baseline_adv}
The baseline and advantage functions $g(S)$ and $f(S)$ are defined as follows:
\begin{align}
    g(S) = [1, S_1, S_2, S_3, S_1 S_2, S_2 S_3, S_1 S_3, S_1 S_2 S_3]\\
    f(S) = [1, S_1, S_2, S_3, S_1 S_2, S_2 S_3, S_1 S_3, S_1 S_2 S_3]
\end{align}

\section{Initialization values and Bayesian Priors}
\label{sec:priors}

\begin{table}[!h]
    \centering
    \begin{tabular}{c|c|c|c}
        Parameter & Significance (S/I/M) & Mean & Variance \\
        \hline
        \hline
        Intercept & S & $2.12$ & $(0.78)^2$\\
        \hline
        $S_1$ & I & $0.00$ & $(0.38)^2$\\
        \hline
        $S_2$ & M & $0.00$ & $(0.62)^2$\\
        \hline
        $S_3$ & S & $-0.69$ & $(0.98)^2$\\
        \hline
        $S_1 S_2$ & M & $0.00$ & $(0.16)^2$\\
        \hline
        $S_1 S_3$ & I & $0.00$ & $(0.1)^2$\\
        \hline
        $S_2 S_3$ & M & $0.00$ & $(0.16)^2$\\
        \hline
        $S_1 S_2 S_3$ & M & $0.00$ & $(0.1)^2$\\
        \hline
        $a$ Intercept & I & $0.00$ & $(0.27)^2$\\
        \hline
        $a S_1$ & I & $0.00$ & ($0.33)^2$\\
        \hline
        $a S_2$ & M & $0.00$ & $(0.3)^2$\\
        \hline
        $a S_3$ & I & $0.00$ & $(0.32)^2$\\
        \hline
        $a S_1 S_2$ & M & $0.00$ & $(0.1)^2$\\
        \hline
        $a S_1 S_3$ & I & $0.00$ & ($0.1)^2$\\
        \hline
        $a S_2 S_3$ & M & $0.00$ & $(0.1)^2$\\
        \hline
        $a S_1 S_2 S_3$ & M & $0.00$ & $(0.1)^2$\\
        \hline
        
    \end{tabular}
    \caption{\textbf{Prior values for the RL algorithm informed using the SARA data set.} The significant column signifies the significant terms found during the analysis using S, insignificant terms using I, and missing terms with M. Values are rounded to the nearest 2 decimal places.}
    \label{tab:informative_prior_vals}
\end{table}

This section details how we calculate the initial values and priors using SARA data (Sec. \ref{sec:testbed}) to estimate the reward, given the state. First, we define the model for the conditional mean of the  reward $R_{i}^{(t)}$ of a given user $i$ at time $t$, given the current state $S = \{S_1, S_2, S_3\}$ (dropping the user index and time for brevity):
\begin{align}
    \notag E\left[R_{i}^{(t)}|S_1, S_2, S_3, a\right] &= \alpha_0 + \alpha_1 S_1 + \alpha_2 S_2 + \alpha_3 S_3 + \alpha_4 S_1 S_2 + \alpha_5 S_1 S_3 + \alpha_6 S_2 S_3 + \alpha_7 S_1 S_2 S_3\\
    &+ a  (\beta_0 + \beta_1 S_1 + \beta_2 S_2 + \beta_3 S_3 + \beta_4 S_1 S_2 + \beta_5 S_1 S_3 + \beta_6 S_2 S_3 + \beta_7 S_1 S_2 S_3).
    \label{eqn:reward_model}
\end{align}

Note that the reward model in Eq. \ref{eqn:reward_model} is non-parametric, i.e. there are 16 unique weights, and $E\left[R_{i}^{(t)}|S_1, S_2, S_3, a\right]$ has 16 dimensions. We follow the methods described in \cite{peng2019} to form our priors using the SARA dataset, which involved fitting linear models like Eq. \ref{eqn:reward_model} using GEE. We do a complete-case analysis on the SARA data, and transform it into State-Action-Reward tuples to mimic our RL algorithm setup. However, as noted in the previous section, the SARA dataset does not account for the entire state-space, specifically $S_2$, i.e. time of day, as the users in the study were requested to self-report on a daily survey just once a day. To that end, we omit all terms from Eq. \ref{eqn:reward_model} which contain $S_2$ when forming our analysis to determine priors. Hence, our reward estimation model while forming priors is given as:
\begin{align}
    E\left[R_{i}^{(t)}|S_1,  S_3, a\right] &= \alpha_0 + \alpha_1 S_1 +  \alpha_3 S_3 +  \alpha_5 S_1 S_3 
    + a  (\beta_0 + \beta_1 S_1 + \beta_3 S_3  + \beta_5 S_1 S_3 ).
    \label{eqn:reward_model_noS2}
\end{align}

\subsection{State formation and imputation}
We operationalize the states of the RL algorithm from the SARA dataset as follows:
\begin{itemize}
    \item $S_1$: Same as defined in Section \ref{sec:rl_framework}
    \item $S_2$: This is not present in the SARA data set.
    \item $S_3$: Same as defined in Section \ref{sec:rl_framework}
\end{itemize}
We work with complete-case data. So, whenever there is missing ``cannabis use'' in the past decision point (since $Y=1$ only), we impute and set $S_3$ to 1. This is because participants in the \trialname study are expected to often use cannabis (at least 3 times a week). Also, since the data is complete-case, we do not use the first day's (i.e. first data point for a user) reward to fit the model, as we choose to not impute the state of the user on the first day while forming our priors.

\subsection{Feature significance}
\begin{figure}[h]
    \centering
    \includegraphics[scale=0.3]{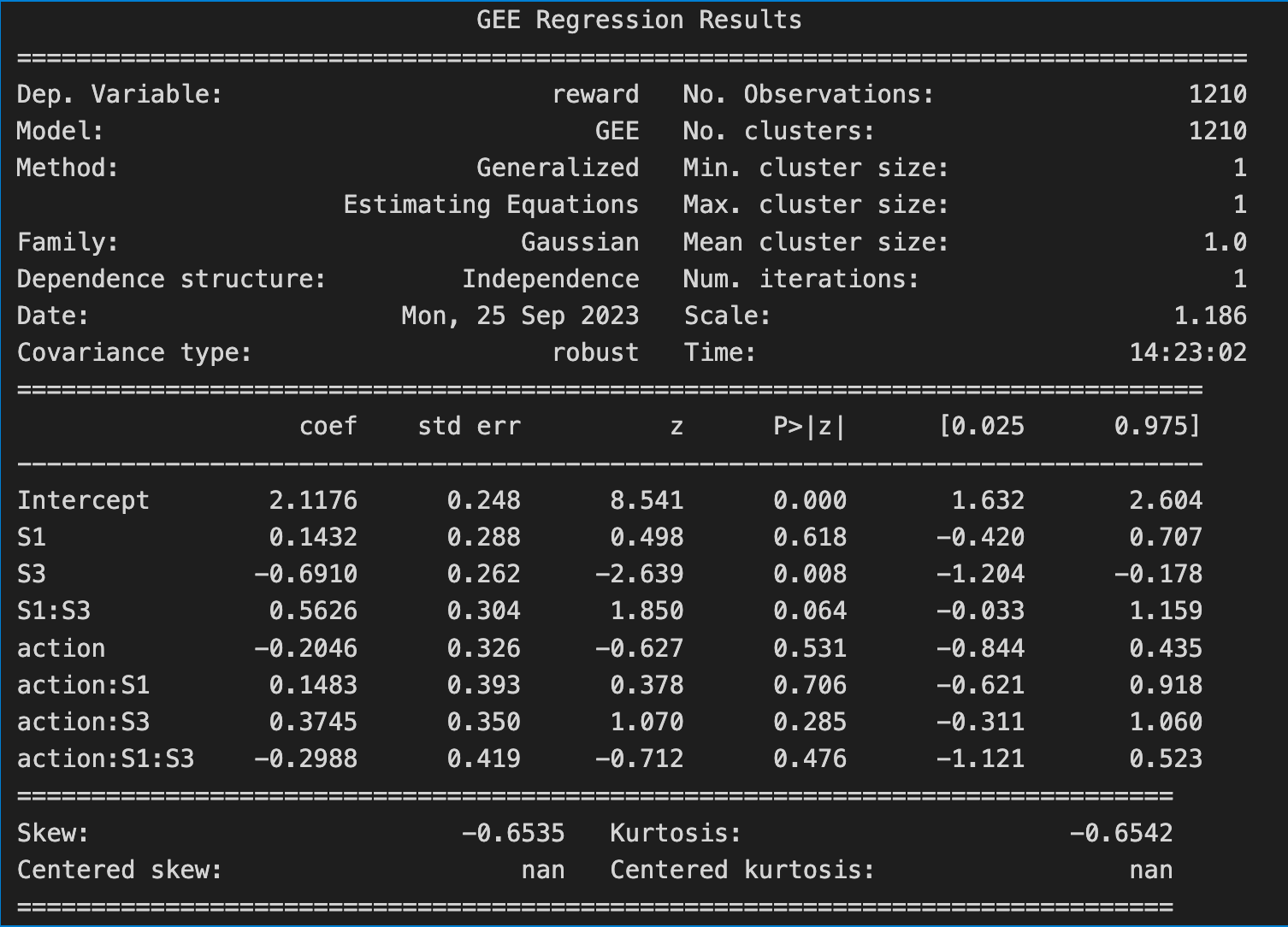}
    \caption{GEE Results}
    \label{fig:gee_analysis}
\end{figure}

We run a GEE linear regression analysis with robust standard errors to determine the feature significance. We categorize a feature to be \emph{significant} when it's corresponding p-value is less than $0.05$. The GEE Regression results are summarized in Fig. \ref{fig:gee_analysis}. Using the criteria mentioned above, we classify the intercept ($\alpha_0$), and the $S_3$ term ($\alpha_3$) to be significant.

\subsection{Initial value of Noise variance (\texorpdfstring{$\sigma^2_{\epsilon}$}{sigma\^2})}
To form an initial of the noise variance, we fit a GEE linear regression model (Eq. \ref{eqn:reward_model_noS2}) per user, and compute the residuals. We set the initial noise variance value as the average of the variance of the residuals across the $N=42$ user models; that is $\sigma^2_{\epsilon, 0} = 0.85$ in our simulations.

\begin{table}[!h]
    \centering
    \begin{tabular}{c|c}
        Parameter & Value \\
        \hline
        \hline
        $\sigma_{\epsilon, 0}^2$ & $0.85$\\
        \hline
        
    \end{tabular}
    \caption{Initial value of noise variance}
    \label{tab:initial_value_noise_var}
\end{table}

\subsection{Prior mean for \texorpdfstring{$\alpha$}{alpha} and \texorpdfstring{$\beta$}{beta}}
To compute the prior means, we first fit a single GEE regression model (Eq. \ref{eqn:reward_model_noS2}) across all the users combined. For the significant features (intercept and $S_3$), we choose the point estimates of $\alpha_0$, and $\alpha_3$ to be their corresponding feature prior means. For the insignificant features, we set the prior mean to 0 ($\alpha_1$, $\alpha_5$, $\beta_0$, $\beta_1$, $\beta_3$, and $\beta_5$). For the prior means of the weights on the $S_2$ terms which are not present in the GEE model, we also set them to 0. 

\subsection{Prior standard deviation for \texorpdfstring{$\alpha$}{alpha} and \texorpdfstring{$\beta$}{beta}}
To compute the prior standard deviation, we first fit user-specific GEE regression models (Equation \ref{eqn:reward_model_noS2}), one per user. We choose the standard deviation of significant features, corresponding to $\alpha_0$, and $\alpha_3$, across the $N=42$ user models to be their corresponding prior standard deviations. For the insignificant features, we choose the standard deviation of $\alpha_1$, $\alpha_5$, $\beta_0$, $\beta_1$, $\beta_3$, and $\beta_5$ across the $N=42$ user models, and set the prior standard deviation to be half of their corresponding values.  The rationale behind setting the mean to 0 and shrinking the prior standard deviation is to ensure stability in the algorithm; we do not expect these terms to play a significant impact on the action selection, unless there is a strong signal in the user data during the trial. In other words, we are reducing the SD of the non-significant weights because we want to provide more shrinkage to the prior mean of 0 (i.e. more data is needed to overcome the prior). For the prior standard deviation of the $S_2$ terms which are in the baseline (i.e. all the $\alpha$ terms), we set it to the average of the prior standard deviations of the other baseline $\alpha$ terms. Similarly, we set the prior standard deviation of the $S_2$ advantage ($\beta$) terms as the average of the prior standard deviations of the other advantage $\beta$ terms. We further shrink the standard deviation of all the two-way interactions by a factor of 4, and the three-way interactions by a factor of 8 - with a minimum prior standard deviation of $0.1$. Recall that we expect little to no signal from the data wrt. the higher order interactions of the binary variables, thus this decision. Unless there is strong evidence in the user data during the \trialname pilot study, we do not expect these interaction terms to play a significant role in action selection. 

\subsection{Initial values for variance of random effects \texorpdfstring{$\Sigma_u$}{Sigma\_u}}
reBandit assumes that each weight $\alpha_{i, j}$ is split into a population term $\alpha_{i, \text{pop}}$ and an individual term $u_{i, j}$ (random effects) for a given feature $i$ and user $j$. These random effects capture the user heterogeneity in the study population, and the variance of the random effects are denoted by $\Sigma_u$. We set the initial values for $\Sigma_u$ to be $\boldsymbol{\Sigma}_{u, 0} = (0.1)^2 \times \boldsymbol{I}_{K}$, where $K$ is the number of random effects terms. We shrink this variance as we allow the data to provide evidence for user heterogeneity. We set the off-diagonal entries in the initial covariance matrix to 0 as we do not have any information about the covariance between the random effects.

\subsection{Empirical check for prior variance shrinkage}
\label{sec:empirical_check}
During on our prior formulation, we shrink the variance of higher order interactions. This shrinkage allows us to deploy complex models without many drawbacks. However, it relies on the idea that if there is evidence of higher order interactions in the study/data, the algorithm will learn and identify that signal. This empirical check is to establish that our algorithms are able to do so, when the study environment does provide evidence of such interactions. In this empirical check, we are primarily concerned with the variance of the action intercept or the advantage intercept coefficient.

\begin{figure}
     \centering
     \begin{subfigure}[b]{0.45\textwidth}
         \centering
         \includegraphics[width=\textwidth]{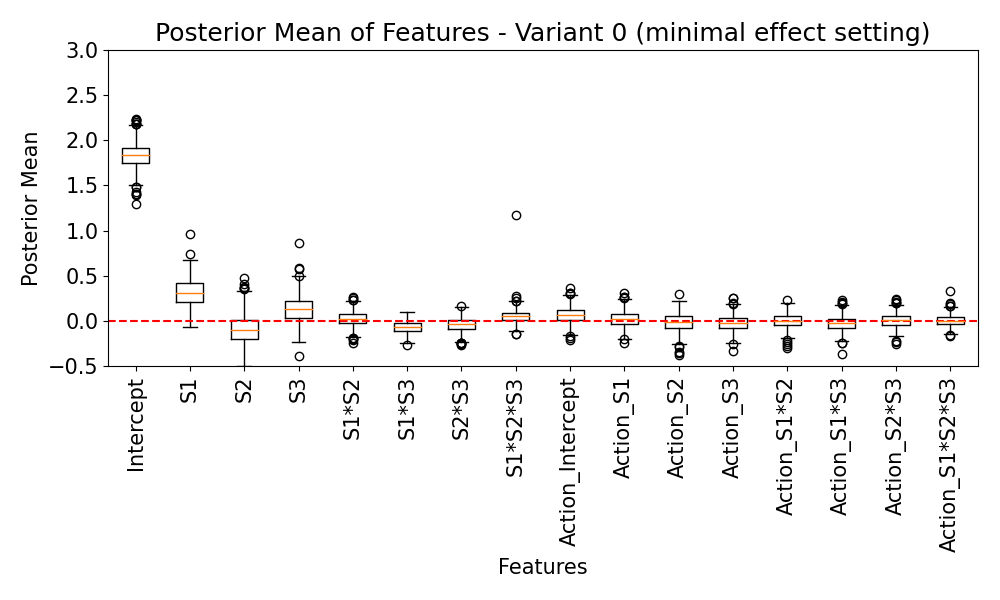}
         \caption{Variant 0: Posterior means}
         \label{fig:post_mean_var0}
     \end{subfigure}
     \hfill
     \begin{subfigure}[b]{0.45\textwidth}
         \centering
         \includegraphics[width=\textwidth]{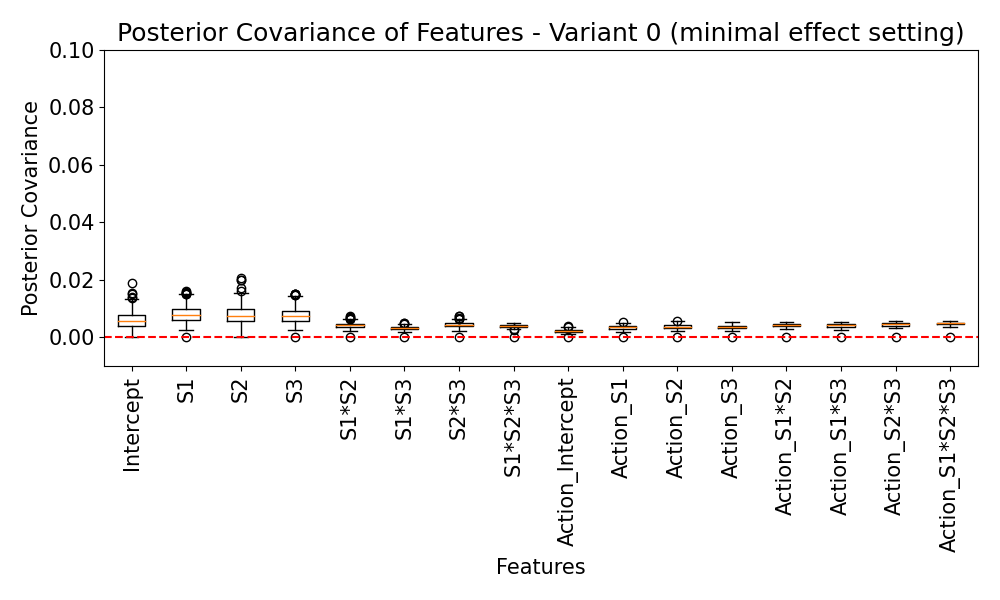}
         \caption{Variant 0: Posterior variance}
         \label{fig:post_cov_var0}
     \end{subfigure}
     \caption{Average posterior means and variances in the minimal treatment effect environment with no habituation}
     \label{fig:posteriors_no_signal_pooled_fixed}
\end{figure}

First, we run experiments with minimal effect/signal. To that end, we run $500$ simulated clinical trials in the minimal treatment effect environment with no habituation, with each trial consisting of $N=120$ simulated users and lasting $T=60$ decision points (30 days). We use smooth posterior sampling to determine the action selection probability (described in Section \ref{sec:act_select}). We also use the BLR algorithm  (described in Appendix \ref{sec:BLR}). Next, we run experiments with evidence/signal in the data, in order to check whether we have shrunk the prior variances too much. To that end, we check if our pooled algorithm is able to identify the signal in the high treatment effect environment with no habituation.

\begin{figure}
     \centering
     \begin{subfigure}[b]{0.45\textwidth}
         \centering
         \includegraphics[width=\textwidth]{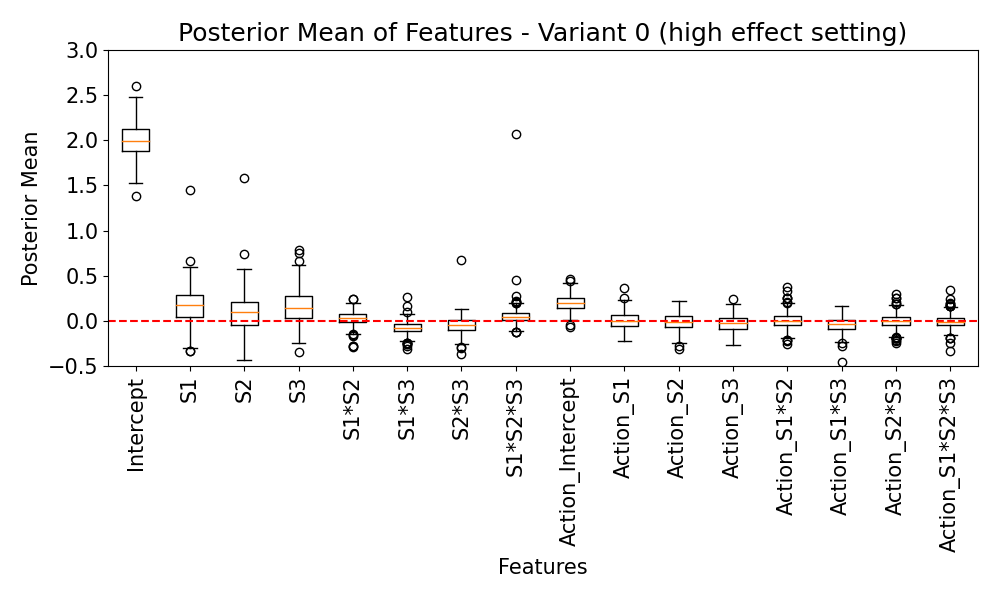}
         \caption{Variant 0: Posterior means}
         \label{fig:post_mean_var0_act_int}
     \end{subfigure}
     \hfill
     \begin{subfigure}[b]{0.45\textwidth}
         \centering
         \includegraphics[width=\textwidth]{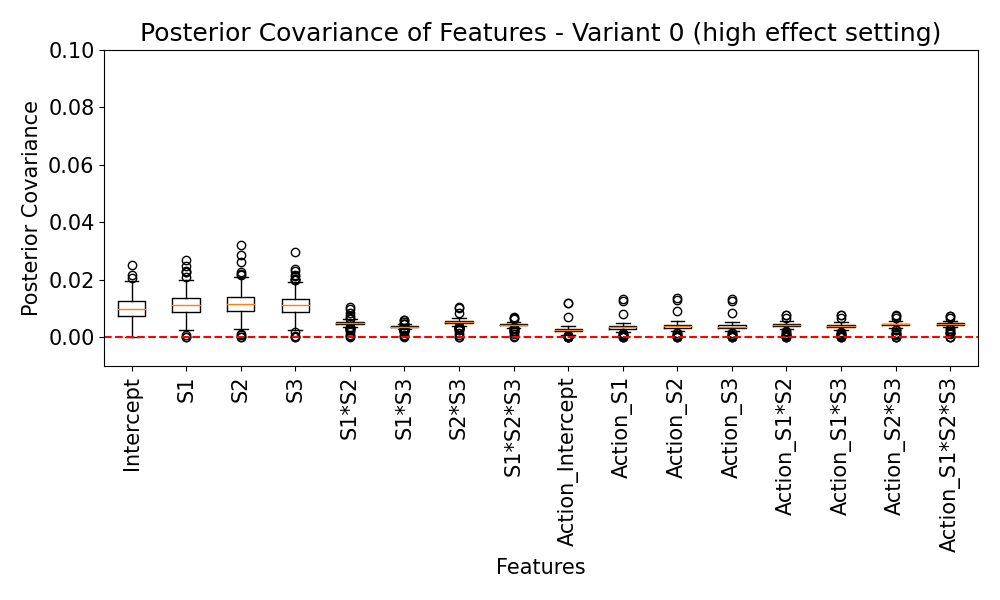}
         \caption{Variant 0: Posterior variance}
         \label{fig:post_cov_var0_act_int}
     \end{subfigure}
     \caption{Average posterior means and variances in high treatment effect environment with no habituation}
     \label{fig:posteriors_act_int_signal_pooled_fixed}
\end{figure}

We compare the three variants in terms of the learnt average posterior means and variances (Figure \ref{fig:posteriors_act_int_signal_pooled_fixed}). We see that the algorithm is able to learn the signal added to the action intercept weight. This demonstrates that the shrinkage of the prior variance of the \emph{action intercept} term is appropriate for learning, when the user data supplies strong evidence of a signal. \bo{Evidence of the algorithm being able to pick up these signals allow us to run more complex models (in terms of the baseline $\bs{g(S)}$ and advantage $\bs{f(S)}$ functions) without many drawbacks}. We are able to do so because we shrink the prior variances of the high order interactions by a lot.

\section{Smooth Posterior Sampling}
\label{sec:smooth_allocation_function}
Recall that the probability of sending an intervention message is computed as:
\begin{align}
    \pi_i^{(t)} = \mathbb{E}_{{\Tilde{\beta} \sim \mathcal{N}(\mu_{\text{post}}^{(t-1)}, \Sigma_{\text{post}}^{(t-1)})}}[\rho(f(S_i^{(t)})^T \boldsymbol{\Tilde{\beta}}) |\mathcal{H}_{1:N}^{(t-1)},  S_{i}^{(t)}]
\end{align}
\begin{equation}
    \rho(x) = L_{\min} + \frac{ L_{\max} - L_{\min} }{  1 + c \exp(-b x) }
    \label{eqn:smooth_post_sampling}
\end{equation}

where $L_{\min} = 0.2$ and $L_{\max} = 0.8$ are the lower and upper clipping probabilities. For now, we set $c=5$. Larger values of $c > 0$ shifts the value of $\rho(0)$ to the right.  This choice implies that $\rho(0)=0.3$. Intuitively, the probability of taking an action when treatment effect is $0$, is $0.3$.

We define $b = \frac{B}{\sigma_{\text{res}}}$. Larger values of $b > 0$ makes the slope of the curve more ``steep''. $\sigma_{\text{res}}$ is the reward residual standard deviation obtained from fitting our reward model on data from SARA. We have $\sigma_{\text{res}}$ in the denominator, as it helps us standardize the treatment effect. The intuition is that the denominator in $\rho(f(s) \beta)$ has the term $\beta / \sigma_{\text{res}}$ in the exponential, which becomes the standardized treatment effect.

We currently set $\sigma_{\text{res}} = 0.95$. In order to arrive at this value, we first simulated each of $N=42$ unique users from SARA in a \trialname simulation with low treatment effect and no habituation effect. The reason we introduce low treatment effect, is because we see that SARA itself had minimal to no treatment effect. Actions are selected with $0.5$ probability. We ran 500 such simulations, and fit each simulation's data into a GEE model, and computed the standard deviation of the residuals. $0.95$ was the mean of the residual standard deviation across these 500 simulations. 

We also set $B=20$. We arrive at this value after comparing the performance of reBandit with $B=10$ and $B=20$ across multiple simulation environments. Hence, we set $b = \frac{B}{\sigma_{\text{res}}} = \frac{20}{0.95} = 21.053$.

\section{Implicit state-features}
\label{sec:induced_states}
The random effects model, described in Sec \ref{sec:reward_model}, induces implicit state-features. The posterior distribution for an individual user can be written in terms of converging population statistics (which use the data across the entire trial population), and the individual user's data (morphed into an implicit feature). While this implicit feature for a user is not apparent from the posterior expressions in Eq. \ref{eqn:postMeanTheta} and Eq. \ref{eqn:postCovTheta}, we derive the joint posterior of the parameters $\tpop{}$ and $\ui{}{i}$ (refer Appendix \ref{sec:implicit_features_derivation}). It turns out that the posterior distribution of 
$\begin{bmatrix}
    \tpop{}\\
    \ui{}{i}
\end{bmatrix}$ is jointly normal:
\begin{align}
    \begin{bmatrix}
        \tpop{}\\
        \ui{}{i}
    \end{bmatrix} \sim \mathcal{N} \bigg(
    \begin{bmatrix}
        \bs{\lambda}\\
        \bs{M}
    \end{bmatrix}, 
    \begin{bmatrix}
        \bs{V_1} & \bs{V_2}\\
        \bs{V_3} & \bs{V_4}
    \end{bmatrix}
    \bigg)
\end{align}

where 
\begin{align}
    \bs{U} &= \psii{-1}{i}(-\A{}{i} \bs{\lambda} + \B{}{i})\\
    \bs{V_1} &= (m\bs{E})^{-1}\\
    \bs{V_2} &= -(m\bs{E})^{-1} \A{}{i} \psii{-1}{i}\\
    \bs{V_3} &= -\psii{-1}{i} \A{}{i} (m\bs{E})^{-1}\\
    \bs{V_4} &= \sige{2}\psii{-1}{i} + \psii{-1}{i}\A{}{i} (m\bs{E})^{-1} \A{}{i} \psii{-1}{i}\\
    \bs{\lambda} &= \bs{E}^{-1}\left(\frac{1}{m}\Sigpinv \muprior + \frac{1}{\sige{2}} \statistic{1} -  \frac{1}{\sige{2}} \statistic{2} \right) \\
    \bs{E} &= \frac{1}{m} \Sigpinv + \frac{1}{\sige{2}} \statistic{3} - \frac{1}{\sige{2}} \statistic{4} \\
    \statistic{1} &= \frac{1}{m} \sum_i \B{}{i}\\
    \statistic{2} &= \frac{1}{m} \sum_{i} \A{}{i} \psii{-1}{i} \B{}{i}\\
    \statistic{3} &= \frac{1}{m} \sum_{i} \A{}{i}\\
    \statistic{4} &= \frac{1}{m} \sum_{i} \A{}{i} \psii{-1}{i} \A{}{i}\\
    \psii{}{i} &= \sige{2} \Sig{-1}{u} + \A{}{i}\\
    \A{}{i} &= \sum_{\tau = 1}^{t} \bs{\Phi}_{i\tau} \bs{\Phi}_{i\tau}^T \label{eqn:Ai}\\
    \B{}{i} &= \sum_{\tau = 1}^{t} \bs{\Phi}_{i\tau} \reward{\tau}{i} \label{eqn:Bi}
\end{align}

We observe that the random effects model induces the state-features $\A{}{i}$ and $\B{}{i}$ described in Eq. \ref{eqn:Ai} and Eq. \ref{eqn:Bi} for each user, which depends on the data for that particular user. It is also evident that the posterior depends on the statistics $\statistic{1}$, $\statistic{2}$, $\statistic{3}$, and $\statistic{4}$ - all of which can be expected to converge in probability as $m$ increases. These implicit features facilitate the after-study analysis of \algorithmname{}, which is left for future work.

\subsection{Derivation}
\label{sec:implicit_features_derivation}

\noindent To start, the mixed effects model can be written as:
\begin{align}
    R_{i}^{(t)} &= \Phii{T}{i,t} \tparam{}{i} + \epsilon_{i, t}\\
    &= \Phii{T}{i,t} (\tpop{} + \ui{}{i}) + \epsilon_{i, t}
\end{align}

And by definition, the random effects are normal with mean 0 and variance $\bs{\Sigma_u}$, i.e. $\bs{u}_i \sim \mathcal{N}(0, \bs{\Sigma_u})$.\\

As described before, we assume the noise to be Gaussian i.e. $\bs{\epsilon} \sim \mathcal{N}(\bs{0}, \sigma_{\epsilon}^2\bs{I}_{mt})$. We also assume the following prior on the population term:
\begin{align}
    \tpop{} \sim \mathcal{N}(\muprior{}, \Sigprior{})
\end{align}

Therefore, we can write the prior distribution as:
\begin{align}
    \log(prior) \propto -\frac{1}{2} (\bs{\tpop{}} - \muprior)^T \bs{\Sig{-1}{\text{prior}}} (\bs{\tpop{}} - \muprior) + \sum_{i=1}^{m} -\frac{1}{2} \bs{u}_i^T \bs{\Sigma^{-1}_u} \bs{u}_i
\end{align}

The likelihood is given as:
\begin{align}
    \log(likelihood) \propto \sum_{i=1}^{m} \sum_{\tau = 1}^{t} - \frac{1}{2\sigma^2_{\epsilon}} (R_{i\tau} - \Phii{T}{i\tau} \bs{\theta}_{\text{pop}} - \Phii{T}{i\tau} \bs{u}_i)^2
\end{align}

Given the prior and the likelihood, the posterior distribution is given as follows:
\begin{align}
    \log(posterior) &= const -\frac{1}{2} (\bs{\tpop{}} - \muprior)^T \bs{\Sig{-1}{\text{prior}}} (\bs{\tpop{}} - \muprior) + \sum_{i=1}^{m} -\frac{1}{2} \bs{u}_i^T \bs{\Sigma^{-1}_u} \bs{u}_i \nonumber\\
    & - \sum_{i=1}^{m} \sum_{\tau = 1}^{t} - \frac{1}{2\sige{2}} (\reward{\tau}{i} - \Phii{T}{i\tau} \bs{\theta}_{\text{pop}} - \Phii{T}{i\tau} \ui{}{i})^2
\end{align}

Define $\bs{A}_i = \sum_{\tau} \bs{\Phi}_{i\tau} \bs{\Phi}_{i\tau}^T$, and $\B{}{i} = \sum_{\tau} \bs{\Phi}_{i\tau} R_{i\tau}$. And,

\begin{align}
    \bs{A} &=
    \begin{bmatrix}
        \sum_{\tau=1}^t \Phii{}{1\tau} \Phii{T}{1\tau} & \bs{0} &\cdots &\bs{0}\\
        \bs{0} & \sum_{\tau=1}^t \Phii{}{2\tau} \Phii{T}{2\tau} & \cdots & \bs{0} \\
        \vdots & \vdots & \ddots & \vdots\\
        \bs{0} & \bs{0} & \cdots & \sum_{\tau=1}^t \Phii{}{m\tau} \Phii{T}{m\tau}
    \end{bmatrix}\\
    \bs{B} &=
    \begin{bmatrix}
        \B{}{1}\\
        \vdots\\
        \B{}{m}\\
    \end{bmatrix} =
    \begin{bmatrix}
        \sum_{\tau=1}^t \Phii{}{1\tau} R^{(\tau)}_{1}\\
        \vdots\\
        \sum_{\tau=1}^t \Phii{}{m\tau} R^{(\tau)}_{m}\\
    \end{bmatrix} 
\end{align}

Then, rewriting the posterior, we get:
\begin{align}
    = const & - \frac{1}{2} (\bs{\theta_{\text{pop}}} - \muprior)^T \Sigpinv (\bs{\theta_{\text{pop}}} - \muprior) - \frac{1}{2} \sum_{i=1}^{m} \bs{u}_i^T \bs{\Sigma^{-1}_u} \bs{u}_i \nonumber \\
    & - \frac{1}{2\sigma^2_{\epsilon}} \bs{\theta}_{\text{pop}}^T \bigg( \sum_{i, \tau} \bs{A}_i \bigg) \bs{\theta}_{\text{pop}} - \frac{1}{2\sigma^2_{\epsilon}} \sum_{i} \bs{u}_{i}^T \bs{A}_i \bs{u}_{i}  \nonumber\\
    &-\frac{1}{\sigma^2_{\epsilon}} \bs{\theta}_{\text{pop}}^T \bigg( \sum_{i} \bs{A}_i \bs{u}_i \bigg) + \frac{1}{\sigma^2_{\epsilon}} \sum_{i} \B{T}{i} \bs{\theta}_{\text{pop}} + \frac{1}{\sigma^2_{\epsilon}} \sum_{i} \B{T}{i} \bs{u}_i \\
    = const & - \frac{1}{2} \bs{\theta^T_{\text{pop}}} \bigg( \Sigpinv + \frac{1}{\sigma_{\epsilon}^2} \sum_{i} \bs{A}_i \bigg) \bs{\theta_{\text{pop}}} + \bigg( \Sigpinv \muprior + \frac{1}{\sigma^2_{\epsilon}} \sum_{i} \B{T}{i} \bigg) \bs{\theta}_{\text{pop}} \nonumber \\
    &-\frac{1}{\sigma^2_{\epsilon}} \theta^T_{\text{pop}} \bs{\tilde{A}} \bs{u} - \frac{1}{2} \bs{u}^T \bs{D} \bs{u} + \frac{1}{\sigma^2_{\epsilon}} \B{T}{} \bs{u}
\end{align}

where $\bs{\tilde{A}} = [\bs{A_1}, \ldots, \bs{A_m}] \in \mathbb{R}^{p \times mp}$, $p=\dim(\design{}{it})$

and
\begin{align}
    \bs{D} &= \begin{bmatrix}
    \Sig{-1}{u} + \frac{1}{\sige{2}} \A{}{1} & \ldots & \bs{0} & \bs{0}\\
    \bs{0} & \Sig{-1}{u} + \frac{1}{\sige{2}}\A{}{2} & \ldots & \bs{0}\\
    \vdots& \vdots & \ddots & \vdots\\
    \bs{0} & \bs{0} & \ldots & \Sig{-1}{u} + \frac{1}{\sige{2}} \A{}{m}
    \end{bmatrix}
\end{align}

Continuing the derivation,

\begin{align}
    \log (posterior) = const & -\frac{1}{2}
    \begin{bmatrix}
        \tpop{}\\
        \ui{}{}
    \end{bmatrix}^T
    \begin{bmatrix}
        \Sigpinv + \frac{1}{\sige{2}} \sum_{i} \A{}{i} & \frac{1}{\sige{2}} \tilde{\A{}{}}\\
        \frac{1}{\sige{2}} \tilde{\A{T}{}} & \bs{D}
    \end{bmatrix}
    \begin{bmatrix}
        \tpop{}\\
        \ui{}{}
    \end{bmatrix}\\
    &+ \begin{bmatrix}
        \Sigpinv \muprior + \frac{1}{\sige{2}} \sum_{i} \B{}{i}\\
        \frac{1}{\sige{2}} \B{}{}
    \end{bmatrix}^T
    \begin{bmatrix}
        \tpop{}\\
        \ui{}{}
    \end{bmatrix}
\end{align}

Continuing the derivation,

\begin{align}
    \log (posterior) = const & -\frac{1}{2}
    \begin{bmatrix}
        \tpop{}\\
        \ui{}{}
    \end{bmatrix}^T
    \begin{bmatrix}
        \Sigpinv + \frac{1}{\sige{2}} \sum_{i} \A{}{i} & \frac{1}{\sige{2}} \tilde{\A{}{}}\\
        \frac{1}{\sige{2}} \tilde{\A{T}{}} & \bs{D}
    \end{bmatrix}
    \begin{bmatrix}
        \tpop{}\\
        \ui{}{}
    \end{bmatrix}\\
    &+ \begin{bmatrix}
        \Sigpinv \muprior + \frac{1}{\sige{2}} \sum_{i} \B{}{i}\\
        \frac{1}{\sige{2}} \B{}{}
    \end{bmatrix}^T
    \begin{bmatrix}
        \tpop{}\\
        \ui{}{}
    \end{bmatrix}
\end{align}

Define
\begin{align}
    \bs{I} &= \begin{bmatrix}
        \Sigpinv + \frac{1}{\sige{2}} \sum_{i} \A{}{i} & \frac{1}{\sige{2}} \tilde{\A{}{}}\\
        \frac{1}{\sige{2}} \tilde{\A{T}{}} & \bs{D}
    \end{bmatrix}\\
    \bs{J}^T &= \begin{bmatrix}
        \Sigpinv \muprior + \frac{1}{\sige{2}} \sum_{i} \B{}{i}\\
        \frac{1}{\sige{2}} \B{}{}
    \end{bmatrix}^T
\end{align}

Then we get,
\begin{align}
    \log (posterior) = const & -\frac{1}{2}
    \begin{bmatrix}
        \tpop{}\\
        \ui{}{}
    \end{bmatrix}^T \bs{I}
    \begin{bmatrix}
        \tpop{}\\
        \ui{}{}
    \end{bmatrix}
    +
    \bs{J}^T
    \begin{bmatrix}
        \tpop{}\\
        \ui{}{}
    \end{bmatrix}\\
    = const & - \frac{1}{2} \Bigg(
    \begin{bmatrix}
        \tpop{}\\
        \ui{}{}
    \end{bmatrix} - \bs{I}^{-1}\bs{J}
    \Bigg)^T
    \bs{I}
    \Bigg(
    \begin{bmatrix}
        \tpop{}\\
        \ui{}{}
    \end{bmatrix} - \bs{I}^{-1}\bs{J}
    \Bigg)
\end{align}

Hence, the joint posterior distribution of $\tpop{}$ and $\ui{}{}$ is jointly normal, as is given as:
\begin{align}
    \begin{bmatrix}
        \tpop{}\\
        \ui{}{}
    \end{bmatrix} \sim 
    \mathcal{N} (\bs{I}^{-1}\bs{J}, \bs{I}^{-1})
\end{align}

Now, we can write, 

\begin{align}
    \bs{I} &= \begin{bmatrix}
        \Sigpinv + \frac{1}{\sige{2}} \sum_{i} \A{}{i} & \frac{1}{\sige{2}} \tilde{\A{}{}}\\
        \frac{1}{\sige{2}} \tilde{\A{T}{}} & \bs{D}
    \end{bmatrix} = 
    \begin{bmatrix}
        \bs{C}_{22} & \bs{C}_{21}\\
        \bs{C}_{12} & \bs{C}_{11}
    \end{bmatrix}\\
    \implies \bs{I}^{-1} &=
    \begin{bmatrix}
        \left(\bs{C}_{22}-\bs{C}_{21} \bs{C}_{11}^{-1} \bs{C}_{12}\right)^{-1} 
        &
        - \left(\bs{C}_{22}-\bs{C}_{21} \bs{C}_{11}^{-1} \bs{C}_{12}\right)^{-1} \bs{C}_{21} \bs{C}_{11}^{-1} \\
	-\bs{C}_{11}^{-1} \bs{C}_{12}\left(\bs{C}_{22}-\bs{C}_{21} \bs{C}_{11}^{-1} \bs{C}_{12}\right)^{-1} & \bs{C}_{11}^{-1}+\bs{C}_{11}^{-1} \bs{C}_{12}\left(\bs{C}_{22}-\bs{C}_{21} \bs{C}_{11}^{-1} \bs{C}_{12} \right)^{-1} \bs{C}_{21} \bs{C}_{11}^{-1}
    \end{bmatrix}\\
    & \triangleq
    \begin{bmatrix}
        (m\bs{E})^{-1} & -(m\bs{E})^{-1} \bs{C}_{21} \bs{C}_{11}^{-1}\\
	-\bs{C}_{11}^{-1} \bs{C}_{12} (m\bs{E})^{-1} & \bs{C}_{11}^{-1}+\bs{C}_{11}^{-1} \bs{C}_{12} (m\bs{E})^{-1} \bs{C}_{21} \bs{C}_{11}^{-1}
    \end{bmatrix}
\end{align}

We evaluate the expression for each block in the block matrix. Then, we get

\begin{align}
    m\bs{E} &= \Sigpinv + \frac{1}{\sige{2}} \sum_{i} \A{}{i} - \frac{1}{\sige{4}} \tilde{\A{}{}} \bs{D} \tilde{\A{}{}}^T\\
    &= \Sigpinv + \frac{1}{\sige{2}} \sum_{i} \A{}{i} - \frac{1}{\sige{4}} \sum_{i} \A{}{i} (\Sig{-1}{u} + \frac{1}{\sige{2}} \A{}{i})^{-1} \A{}{i}\\
    &= \Sigpinv + \frac{1}{\sige{2}} \sum_{i} \A{}{i} - \frac{1}{\sige{2}} \sum_{i} \A{}{i} \psii{-1}{i} \A{}{i}\\
    \implies \bs{E} &= \frac{1}{m}\bigg( \Sigpinv + \frac{1}{\sige{2}} \sum_{i} \A{}{i} - \frac{1}{\sige{2}} \sum_{i} \A{}{i} \psii{-1}{i} \A{}{i} \bigg)
\end{align}

where $\psii{}{i} = \sige{2} \Sig{-1}{u} + \A{}{i}$.\\


Next,
\begin{align}
    -\bs{C}_{11}^{-1} \bs{C}_{12} (m\bs{E})^{-1} &= -\frac{1}{\sige{2}} \bs{D}^{-1}  \tilde{\A{}{}}^T (m\bs{E})^{-1}\\
    & =-
    \begin{bmatrix}
        \psii{-1}{1}\A{}{1}\\
        \vdots\\
        \psii{-1}{m}\A{}{m}
    \end{bmatrix} (m\bs{E})^{-1}
\end{align}

Next,
\begin{align}
    -(m\bs{E})^{-1}\bs{C}_{21}\bs{C}_{11}^{-1} 
    & =- (m\bs{E})^{-1}
    \begin{bmatrix}
        \A{}{1} \psii{-1}{1}&
        \ldots &
        \A{}{m} \psii{-1}{m}
    \end{bmatrix} 
\end{align}

And finally,
\begin{align}
    \bs{C}_{11}^{-1}+\bs{C}_{11}^{-1} \bs{C}_{12} (m\bs{E})^{-1} \bs{C}_{21} \bs{C}_{11}^{-1} &= 
    \bs{D}^{-1} + \Bigg[\psii{-1}{i}\A{}{i} (m\bs{E})^{-1} \A{}{j} \psii{-1}{j}\Bigg]_{i, j  \in [m]}
\end{align}

Plugging in these values, we get
\begin{align}
    \bs{I}^{-1}\bs{J} &= 
    \begin{bmatrix}
        (m\bs{E})^{-1} & -(m\bs{E})^{-1} \bs{C}_{21} \bs{C}_{11}^{-1}\\
    	-\bs{C}_{11}^{-1} \bs{C}_{12} (m\bs{E})^{-1} & \bs{C}_{11}^{-1}+\bs{C}_{11}^{-1} \bs{C}_{12} (m\bs{E})^{-1} \bs{C}_{21} \bs{C}_{11}^{-1}
    \end{bmatrix}
    \begin{bmatrix}
        \Sigpinv \muprior+\frac{1}{\sige{2}} \sum_i \B{}{i} \\
        \frac{1}{\sige{2}} \B{}{}
    \end{bmatrix}\\
    & = 
    \begin{bmatrix}
        (m\bs{E})^{-1} (\Sigpinv \muprior+\frac{1}{\sige{2}} \sum_i \B{}{i}) - \frac{1}{\sige{2}} (m\bs{E})^{-1} \bs{C}_{21} \bs{C}_{11}^{-1} \B{}{}\\
        -\bs{C}_{11}^{-1} \bs{C}_{12} (m\bs{E})^{-1} (\Sigpinv \muprior+\frac{1}{\sige{2}} \sum_i \B{}{i}) +
        \frac{1}{\sige{2}} (\bs{C}_{11}^{-1}+\bs{C}_{11}^{-1} \bs{C}_{12} (m\bs{E})^{-1} \bs{C}_{21} \bs{C}_{11}^{-1}) \B{}{}
    \end{bmatrix}
    \label{eqn:post_mean_joint_blockmatrix_app}
\end{align}

Then, the upper block matrix evaluates to:
\begin{align}
    & (m\bs{E})^{-1} (\Sigpinv \muprior+\frac{1}{\sige{2}} \sum_i \B{}{i}) - \frac{1}{\sige{2}} (m\bs{E})^{-1} \bs{C}_{21} \bs{C}_{11}^{-1} \B{}{}\\\
    & = (m\bs{E})^{-1} (\Sigpinv \muprior+\frac{1}{\sige{2}} \sum_i \B{}{i}) - \frac{1}{\sige{2}} (m\bs{E})^{-1}
    \begin{bmatrix}
        \A{}{1} \psii{-1}{1}&
        \ldots &
        \A{}{m} \psii{-1}{m}
    \end{bmatrix} \B{}{}\\
    &= (m\bs{E})^{-1} (\Sigpinv \muprior+\frac{1}{\sige{2}} \sum_i \B{}{i}) - \frac{1}{\sige{2}} (m\bs{E})^{-1} \sum_{i} \A{}{i} \psii{-1}{i} \B{}{i}\\
    &= (\bs{E})^{-1} \frac{1}{m}(\Sigpinv \muprior+\frac{1}{\sige{2}} \sum_i \B{}{i}) - \frac{1}{\sige{2}} (\bs{E})^{-1} \frac{1}{m}\sum_{i} \A{}{i} \psii{-1}{i} \B{}{i}
    \label{eqn:post_mean_joint_fixed_app}
\end{align}


Next, the bottom term in the block matrix of \ref{eqn:post_mean_joint_blockmatrix_app} evaluates to:
\begin{align}
    &-\bs{C}_{11}^{-1} \bs{C}_{12} \underbrace{(m\bs{E})^{-1}\left(\Sigpinv \muprior + \frac{1}{\sige{2}} \sum_i \B{}{i}\right)}_{\triangleq \bs{f}} + \frac{1}{\sige{2}}\left(\bs{C}_{11}^{-1}+\bs{C}_{11}^{-1} \bs{C}_{22} (m\bs{E})^{-1} \bs{C}_{21} \bs{C}_{11}^{-1}\right) \B{}{} \\
    &= -\bs{C}_{11}^{-1} \bs{C}_{12} \bs{f}+\frac{1}{\sige{2}} \bs{D}^{-1} \B{}{} +
    \frac{1}{\sige{2}} \Bigg[\psii{-1}{i}\A{}{i} (m\bs{E})^{-1} \A{}{j} \psii{-1}{j}\Bigg]_{i, j  \in [m]} \B{}{}\\
    &= -
    \begin{bmatrix}
        \psii{-1}{1}\A{}{1}\\
        \vdots\\
        \psii{-1}{m}\A{}{m}
    \end{bmatrix} \bs{f}
    +
    \begin{bmatrix}
        \psii{-1}{1} \B{}{1}\\
        \vdots\\
        \psii{-1}{m} \B{}{m}
    \end{bmatrix} 
    +
    \begin{bmatrix}
        \psii{-1}{1}\A{}{1}\\
        \vdots\\
        \psii{-1}{m}\A{}{m}
    \end{bmatrix}
    \underbrace{\frac{1}{\sige{2}} (m\bs{E})^{-1} \sum_{i} \A{}{i} \psii{-1}{i} \B{}{i}}_{\triangleq \bs{g}}\\
    &=
    \begin{bmatrix}
        \psii{-1}{1}\A{}{1}\\
        \vdots\\
        \psii{-1}{m}\A{}{m}
    \end{bmatrix} (\bs{g} - \bs{f})
    + 
    \begin{bmatrix}
        \psii{-1}{1} \B{}{1}\\
        \vdots\\
        \psii{-1}{m} \B{}{m}
    \end{bmatrix}
\end{align}


Thus, posterior distribution of 
$\begin{bmatrix}
    \tpop{}\\
    \ui{}{i}
\end{bmatrix}$ is jointly normal with posterior mean:
\begin{align}
    \begin{bmatrix}
        \tpop{}\\
        \ui{}{i}
    \end{bmatrix} \sim \mathcal{N} \bigg(
    \begin{bmatrix}
        \bs{\lambda}\\
        \psii{-1}{i}(-\A{}{i} \bs{\lambda} + \B{}{i})
    \end{bmatrix}, 
    \begin{bmatrix}
        (m\bs{E})^{-1} & -(m\bs{E})^{-1} \A{}{i} \psii{-1}{i}\\
         -\psii{-1}{i} \A{}{i} (m\bs{E})^{-1} & 
         \sige{2}\psii{-1}{i} + \psii{-1}{i}\A{}{i} (m\bs{E})^{-1} \A{}{i} \psii{-1}{i}
    \end{bmatrix}
    \bigg)
\end{align}

where 
\begin{align}
    \bs{\lambda} &= \bs{E}^{-1}\left(\frac{1}{m}\Sigpinv \muprior + \frac{1}{\sige{2}} \statistic{1} -  \frac{1}{\sige{2}} \statistic{2}  \right)\\
    \bs{E} &= \frac{1}{m} \Sigpinv + \frac{1}{\sige{2}} \statistic{3} - \frac{1}{\sige{2}} \statistic{4} \\
    \statistic{1} &= \frac{1}{m} \sum_i \B{}{i}\\
    \statistic{2} &= \frac{1}{m} \sum_{i} \A{}{i} \psii{-1}{i} \B{}{i}\\
    \statistic{3} &= \frac{1}{m} \sum_{i} \A{}{i}\\
    \statistic{4} &= \frac{1}{m} \sum_{i} \A{}{i} \psii{-1}{i} \A{}{i}\\
    \psii{}{i} &= \sige{2} \Sig{-1}{u} + \A{}{i}\\
    \A{}{i} &= \sum_{\tau} \bs{\Phi}_{i\tau} \bs{\Phi}_{i\tau}^T \label{eqn:Ai_app}\\
    \B{}{i} &= \sum_{\tau} \bs{\Phi}_{i\tau} \reward{\tau}{i} \label{eqn:Bi_app}
\end{align}

\section{Posterior update derivation}
\label{appendix:posterior}
\allowdisplaybreaks
The posterior distribution can be derived as follows. Given the following,

\begin{align}
    \tparam{}{} &= \tpop{} + \ui{}{}\\
    \tpop{} &\sim \mathcal{N}(\muprior, \Sigprior)\\
    \ui{}{i} &\sim \mathcal{N}(\bs{0}, \bs{\Sigma}_{u})\\
    \implies \bs{\theta} &\sim \mathcal{N}(\bs{\mu_{\theta}}, \Sig{}{\theta})
\end{align}

where

\begin{align}
    \bs{\mu_{\theta}} &= 
    \begin{bmatrix}
        \bs{\muprior} \\
        \bs{\muprior} \\
        \vdots\\
        \bs{\muprior}
    \end{bmatrix} \label{eqn:mu_t0_app}\\
    \Sig{}{\theta} &= 
    \begin{bmatrix}
        \bs{\Sigprior} + \bs{\Sigma_{u}} & \bs{\Sigprior} & \cdots & \bs{\Sigprior}\\
        \bs{\Sigprior} & \bs{\Sigprior} + \bs{\Sigma_{u}} & \cdots & \bs{\Sigprior}\\
        \vdots & \vdots & \ddots & \vdots\\
        \bs{\Sigprior} & \bs{\Sigprior} & \cdots & \bs{\Sigprior} + \bs{\Sigma_{u}}
    \end{bmatrix} \label{eqn:Sig_t0_app}
\end{align}

Here $\dim(\muprior) = p$, $\dim(\Sigprior) = p \times p$, $\dim(\bs{\mu_{\theta}}) = pm$ and $\dim(\Sig{}{\theta}) = pm \times pm$, where $p = \dim(\Phii{}{})$. Then, define

\begin{align}
    \bs{\Tilde{\Sigma}_{\theta, t}} &= 
    \begin{bmatrix}
        \bs{\Sigprior} + \bs{\Sigma_{u,t}} & \bs{\Sigprior} & \cdots & \bs{\Sigprior}\\
        \bs{\Sigprior} & \bs{\Sigprior} + \bs{\Sigma_{u,t}} & \cdots & \bs{\Sigprior}\\
        \vdots & \vdots & \ddots & \vdots\\
        \bs{\Sigprior} & \bs{\Sigprior} & \cdots & \bs{\Sigprior} + \bs{\Sigma_{u,t}}
    \end{bmatrix}
\end{align}

So, we can write:
\begin{align}
    prior &\propto \exp\bigg(-\frac{1}{2} (\bs{\theta} - \bs{\mu_{\theta}})^T \bs{\Tilde{\Sigma}^{-1}_{\theta, t}} (\bs{\theta} - \bs{\mu_{\theta}}) \bigg)\\
    likelihood &\propto \prod_{i, \tau} \exp \bigg(-\frac{1}{2 \siget{2}} (R^{(\tau)}_{i} - \Phii{T}{i,\tau} \bs{\theta_i})^2 \bigg)\\
    posterior &\propto prior \times likelihood\\
    \implies log(posterior) &\propto -\frac{1}{2} (\bs{\theta} - \bs{\mu_{\theta}})^T \bs{\Tilde{\Sigma}^{-1}_{\theta, t}} (\bs{\theta} - \bs{\mu_{\theta}}) -\frac{1}{2 \siget{2}} \sum_{i=1}^m \sum_{\tau=1}^t (R^{(\tau)}_{i} - \Phii{T}{i,\tau} \bs{\theta_i})^2\\
    &\propto -\frac{1}{2} (\bs{\theta} - \bs{\mu_{\theta}})^T \bs{\Tilde{\Sigma}^{-1}_{\theta, t}} (\bs{\theta} - \bs{\mu_{\theta}}) -\frac{1}{2 \siget{2}} \sum_{i=1}^m (\bs{\theta_i}^T (\sum_{\tau=1}^t \Phii{}{i \tau} \Phii{T}{i,\tau} )\bs{\theta_i}  \nonumber \\
    &- 2 (\sum_{\tau=1}^n \Phii{}{i \tau} R^{(\tau)}_{i})^T \bs{\theta_i})  
\end{align}

Now, we can define:
\begin{align}
    \bs{A} &=
    \begin{bmatrix}
        \sum_{\tau=1}^t \Phii{}{1\tau} \Phii{T}{1\tau} & \bs{0} &\cdots &\bs{0}\\
        \bs{0} & \sum_{\tau=1}^t \Phii{}{2\tau} \Phii{T}{2\tau} & \cdots & \bs{0} \\
        \vdots & \vdots & \ddots & \vdots\\
        \bs{0} & \bs{0} & \cdots & \sum_{\tau=1}^t \Phii{}{m\tau} \Phii{T}{m\tau}
    \end{bmatrix}\\
    \bs{B} &=
    \begin{bmatrix}
        \sum_{\tau=1}^t \Phii{}{1\tau} R^{(\tau)}_{1}\\
        \vdots\\
        \sum_{\tau=1}^t \Phii{}{m\tau} R^{(\tau)}_{m}\\
    \end{bmatrix}
\end{align}

Then, we can rewrite the log posterior as:
\begin{align}
    log(posterior) &\propto -\frac{1}{2} \tparam{T}{} \bs{\Tilde{\Sigma}^{-1}_{\theta, t}} \bs{\theta} + (\bs{\Tilde{\Sigma}^{-1}_{\theta, t}} \bs{\mu_{\theta}})^T \bs{\theta} -\frac{1}{2 \siget{2}} \tparam{T}{} \bs{A} \bs{\theta} + \frac{1}{\siget{2}} \bs{B}^T \bs{\theta}\\
    &= - \frac{1}{2} \tparam{T}{} \bigg(\bs{\Tilde{\Sigma}^{-1}_{\theta, t}} + \frac{1}{\siget{2}} \bs{A} \bigg)\bs{\theta} + \bigg( (\bs{\Tilde{\Sigma}^{-1}_{\theta, t}} \bs{\mu_{\theta}})^T + \frac{1}{\siget{2}} \bs{B}^T\bigg) \bs{\theta}
\end{align}

Therefore, we can write the posterior mean and variance of the parameters $\bs{\theta}$ at time $t$ as:
\begin{align}
    \bs{\mu_{\text{post}, t}} &= \bigg(\bs{\Tilde{\Sigma}^{-1}_{\theta, t}} + \frac{1}{\siget{2}}\bs{A} \bigg)^{-1} \bigg( \bs{\Tilde{\Sigma}^{-1}_{\theta, t}} \bs{\mu_{\theta}} + \frac{1}{\siget{2}}\bs{B} \bigg) \label{eqn:postMeanTheta_app} \\
    \Sig{}{\text{post}, t} &= \bigg(\bs{\Tilde{\Sigma}^{-1}_{\theta, t}} + \frac{1}{\siget{2}} \bs{A} \bigg)^{-1} \label{eqn:postCovTheta_app}
\end{align}

\section{Hyper-parameter update derivation}
\label{appendix:hyperparam}
Using Empirical Bayes to maximize the marginal likelihood of observed rewards, marginalized over the parameters $\tparam{}{}$, we get:
\begin{align}
    l(\Sig{}{u}, \siget{2} ; \mathcal{H}) = \int_{\bs{\theta}} \prod_{\tau \in [t]} \prod_{i \in [m]} l(R^{(\tau)}_{i} | \bs{\theta_i}) \Pr(\bs{\theta} | \Sig{}{u}, \siget{2}) d\bs{\theta}
\end{align}

Now,
\begin{align}
    \Pr(\bs{\theta} | \Sig{}{u}, \siget{2}) d\bs{\theta} &= \Pr(\bs{\theta} | \Sig{}{u}) d\bs{\theta}\\
    &= \frac{1}{\sqrt{(2 \pi)^{2m} \det(\bs{\Tilde{\Sigma}_{\theta, t}})}} \exp\bigg({-\frac{1}{2}(\bs{\theta} - \bs{\mu_{\theta}})^T } \bs{\Tilde{\Sigma}^{-1}_{\theta, t}} (\bs{\theta} - \bs{\mu_{\theta}})\bigg)
\end{align}
And,
\begin{align}
    \prod_{\tau \in [t]} \prod_{i \in [m]} l(R^{(\tau)}_{i} | \bs{\theta_i}) &= \prod_{\tau \in [t]} \prod_{i \in [m]} \frac{1}{\sqrt{2 \pi \siget{2}}} \exp \bigg({-\frac{1}{2 \siget{2}}} (R^{(\tau)}_{i} - \Phii{T}{i\tau} \bs{\theta}_i)^2 \bigg)\\
    &= \frac{1}{(2 \pi \siget{2}) ^{\frac{mt}{2}}} \exp \bigg({-\frac{1}{2 \siget{2}}} (\tparam{T}{} \bs{A} \bs{\theta} - 2 \bs{B}^T \bs{\theta} + \sum_{\tau\in [t]} \sum_{i\in [m]} (R^{(\tau)}_{i})^2) \bigg)
\end{align}

Therefore, we can write:
\allowdisplaybreaks
\begin{align}
    l(\Sig{}{u}, \siget{2} ; \mathcal{H}) &= \int_{\bs{\theta}} \frac{1}{\sqrt{(2 \pi)^{2m} \det(\bs{\Tilde{\Sigma}_{\theta, t}})}} \exp\bigg({-\frac{1}{2}(\bs{\theta} - \bs{\mu_{\theta}})^T } \bs{\Tilde{\Sigma}^{-1}_{\theta, t}} (\bs{\theta} - \bs{\mu_{\theta}})\bigg) \nonumber\\
    & \frac{1}{(2 \pi \siget{2}) ^{\frac{mt}{2}}} \exp \bigg(-\frac{1}{2 \siget{2}} (\tparam{T}{} \bs{A} \bs{\theta} - 2 \bs{B}^T \bs{\theta} + \sum_{\tau\in [t]} \sum_{i\in [m]} (R^{(\tau)}_{i})^2) \bigg) d\bs{\theta}\\
    &= \int_{\bs{\theta}}\;\;\;\; \overbrace{\bigg (\frac{1}{\sqrt{(2 \pi)^{2m} \det(\bs{\Tilde{\Sigma}_{\theta, t}})}} \bigg) \cdot
    \bigg( \frac{1}{(2 \pi \siget{2}) ^{\frac{mt}{2}}} \bigg) \cdot \exp \bigg( {-\frac{1}{2 \siget{2}} \sum_{\tau\in [t]} \sum_{i\in [m]} (R^{(\tau)}_{i})^2}  \bigg)}^{C_1} \nonumber\\
    & \exp \bigg( {-\frac{1}{2}(\bs{\theta} - \bs{\mu_{\theta}})^T } \bs{\Tilde{\Sigma}^{-1}_{\theta, t}} (\bs{\theta} - \bs{\mu_{\theta}}) -\frac{1}{2 \siget{2}} (\tparam{T}{} \bs{A} \bs{\theta} - 2 \bs{B}^T \bs{\theta})\bigg)  d\bs{\theta}\\
    &= C_1 \int_{\bs{\theta}} \;\; \overbrace{\exp \bigg(\frac{1}{2} (\bs{\Tilde{\Sigma}^{-1}_{\theta, t}} \bs{\mu_{\theta}}  + \frac{1}{\siget{2}} \bs{B})^T (\bs{\Tilde{\Sigma}^{-1}_{\theta, t}} + \frac{1}{\siget{2}} \bs{A} )^{-1} (\bs{\Tilde{\Sigma}^{-1}_{\theta, t}} \bs{\mu_{\theta}}  + \frac{1}{\siget{2}} \bs{B}) - \frac{1}{2} \bs{\mu_{\theta}^T} \bs{\Tilde{\Sigma}^{-1}_{\theta, t}} \bs{\mu_{\theta}}\bigg)}^{C_2} \nonumber\\
    & \exp \bigg( -\frac{1}{2} \bigg(\bs{\theta} - \bigg(\bs{\Tilde{\Sigma}^{-1}_{\theta, t}} + \frac{1}{\siget{2}}\bs{A} \bigg)^{-1} \bigg( \bs{\Tilde{\Sigma}^{-1}_{\theta, t}} \bs{\mu_{\theta}} + \frac{1}{\siget{2}}\bs{B} \bigg) \bigg)^T \nonumber \\
    & \bigg( \bigg(\bs{\Tilde{\Sigma}^{-1}_{\theta, t}} + \frac{1}{\siget{2}} \bs{A} \bigg)^{-1} \bigg) ^{-1}
    \bigg(\bs{\theta} - \bigg(\bs{\Tilde{\Sigma}^{-1}_{\theta, t}} + \frac{1}{\siget{2}}\bs{A} \bigg)^{-1} \bigg( \bs{\Tilde{\Sigma}^{-1}_{\theta, t}} \bs{\mu_{\theta}} + \frac{1}{\siget{2}}\bs{B} \bigg) \bigg) \bigg)  d\bs{\theta}\\
    &= C_1 C_2 \sqrt{(2\pi)^{2m} \det\bigg( \bigg(\bs{\Tilde{\Sigma}^{-1}_{\theta, t}} + \frac{1}{\siget{2}} \bs{A} \bigg)^{-1} \bigg)}\\
    &= \Bigg(\frac{1}{\sqrt{(2\pi)^{mt}\det \bigg(\bs{I} + \frac{1}{\siget{2}} \bs{\Tilde{\Sigma}_{\theta, t}} \bs{A} \bigg)}} \Bigg) \Bigg(\frac{1}{(\siget{2}) ^{\frac{mt}{2}}}\Bigg) \exp \bigg({-\frac{1}{2 \siget{2}} \sum_{\tau\in [t]} \sum_{i\in [m]} (R^{(\tau)}_{i})^2} - \frac{1}{2} \bs{\mu_{\theta}^T} \bs{\Tilde{\Sigma}^{-1}_{\theta, t}} \bs{\mu_{\theta}} \bigg) \nonumber \\
    &  \exp \bigg( \frac{1}{2} (\bs{\Tilde{\Sigma}^{-1}_{\theta, t}} \bs{\mu_{\theta}}  + \frac{1}{\siget{2}} \bs{B})^T (\bs{\Tilde{\Sigma}^{-1}_{\theta, t}} + \frac{1}{\siget{2}} \bs{A} )^{-1} (\bs{\Tilde{\Sigma}^{-1}_{\theta, t}} \bs{\mu_{\theta}}  + \frac{1}{\siget{2}} \bs{B})\bigg) \label{eqn:lhood_eb_app}
\end{align}
\begin{align}
    \implies \log(l(\Sig{}{u}, \siget{2} ; \mathcal{H})) &\propto \log(\det(\bs{X})) - \log(\det(\bs{X} + y \bs{A})) + mt \log(y) - y \sum_{\tau\in [t]} \sum_{i \in [m]} (R^{(\tau)}_{i})^2 \nonumber \\
    & - \bs{\mu_{\theta}^T} \bs{X} \bs{\mu_{\theta}} + (\bs{X} \bs{\mu_{\theta}}  + y \bs{B})^T (\bs{X} + y \bs{A} )^{-1} (\bs{X} \bs{\mu_{\theta}}  + y \bs{B})
\end{align}
where $\bs{X} = \bs{\Tilde{\Sigma}^{-1}_{\theta, t}}$ and $y = \frac{1}{\sigma_{\epsilon, t}^2}$.

\end{document}